\pdfoutput=1

\documentclass[11pt]{article}

\usepackage{ACL2023}

\usepackage{times}
\usepackage{latexsym}
\usepackage{enumitem}

\usepackage[T1]{fontenc}

\usepackage[utf8]{inputenc}

\usepackage{microtype}

\usepackage{inconsolata}

\usepackage{graphicx} 
\usepackage{bm}
\usepackage{tabularray}
\usepackage{booktabs}
\usepackage{color}
\usepackage{diagbox}
\usepackage{multirow}
\usepackage{float}
\usepackage{stfloats}
\usepackage{amsmath}
\usepackage{framed}
\usepackage{amsfonts}
\usepackage{xurl}
\usepackage{colortbl}
\usepackage[normalem]{ulem}
\usepackage{lipsum}
\usepackage{subfigure} 
\usepackage{comment}
\excludecomment{mycomment}
\usepackage{makecell}
\usepackage{tabularx}

\title{Detecting Knowledge Boundary of Vision Large Language Models by Sampling-Based Inference}

\author{Zhuo Chen$^{1}$, Xinyu Wang$^{2}$\thanks{\hspace{0.3em} Corresponding author}, Yong Jiang$^{2*}$, Zhen Zhang$^{2}$, Xinyu Geng$^{2}$, \\ {\bf Pengjun Xie$^{2}$, Fei Huang$^{2}$, Kewei Tu$^{1*}$} \\
        $^{1}$School of Information Science and Technology, ShanghaiTech University \\
        $^{1}$Shanghai Engineering Research Center of Intelligent Vision and Imaging \\
        $^{2}$Institute for Intelligent Computing, Alibaba Group \\
        \texttt{chenzhuo@shanghaitech.edu.cn} \\
}

\begin{document}

\maketitle

\begin{abstract}
Despite the advancements made in Vision Large Language Models (VLLMs), like text Large Language Models (LLMs), they have limitations in addressing questions that require real-time information or are knowledge-intensive. 
Indiscriminately adopting Retrieval Augmented Generation (RAG) techniques is an effective yet expensive way to enable models to answer queries beyond their knowledge scopes. To mitigate the dependence on retrieval and simultaneously maintain, or even improve, the performance benefits provided by retrieval, we propose a method to detect the knowledge boundary of VLLMs, allowing for more efficient use of techniques like RAG. 
Specifically, we propose a method with two variants that fine-tune a VLLM on an automatically constructed dataset for boundary identification. 
Experimental results on various types of Visual Question Answering datasets show that our method successfully depicts a VLLM's knowledge boundary, based on which we are able to reduce indiscriminate retrieval while maintaining or improving the performance. 
In addition, we show that the knowledge boundary identified by our method for one VLLM can be used as a surrogate boundary for other VLLMs.
Code will be released at \url{https://github.com/Chord-Chen-30/VLLM-KnowledgeBoundary}
\end{abstract}

\section{Introduction}

The great advancements in language models have led to the integration of image encoding and understanding capabilities \cite{achiam2023gpt,lu2024deepseekvl, Qwen2-VL}, significantly enhancing the performance of a series of pre-trained  Vision Large Language Models (VLLMs) in tasks involving Visual Question Answering (VQA). Despite these advancements, akin to Large Language Models (LLMs) \cite{touvron2023llama, workshop2022bloom, brown2020language, zhang2024exploring}, VLLMs remain constrained by the boundaries of their knowledge \cite{lin-byrne-2022-retrieval}. As a result, their ability to accurately respond to content outside the model's knowledge scope, such as knowledge-intensive questions, real-time news, and queries with dynamic answers, is considerably limited. 

\begin{figure}[tb]
\centering
\includegraphics[width=0.87\linewidth]{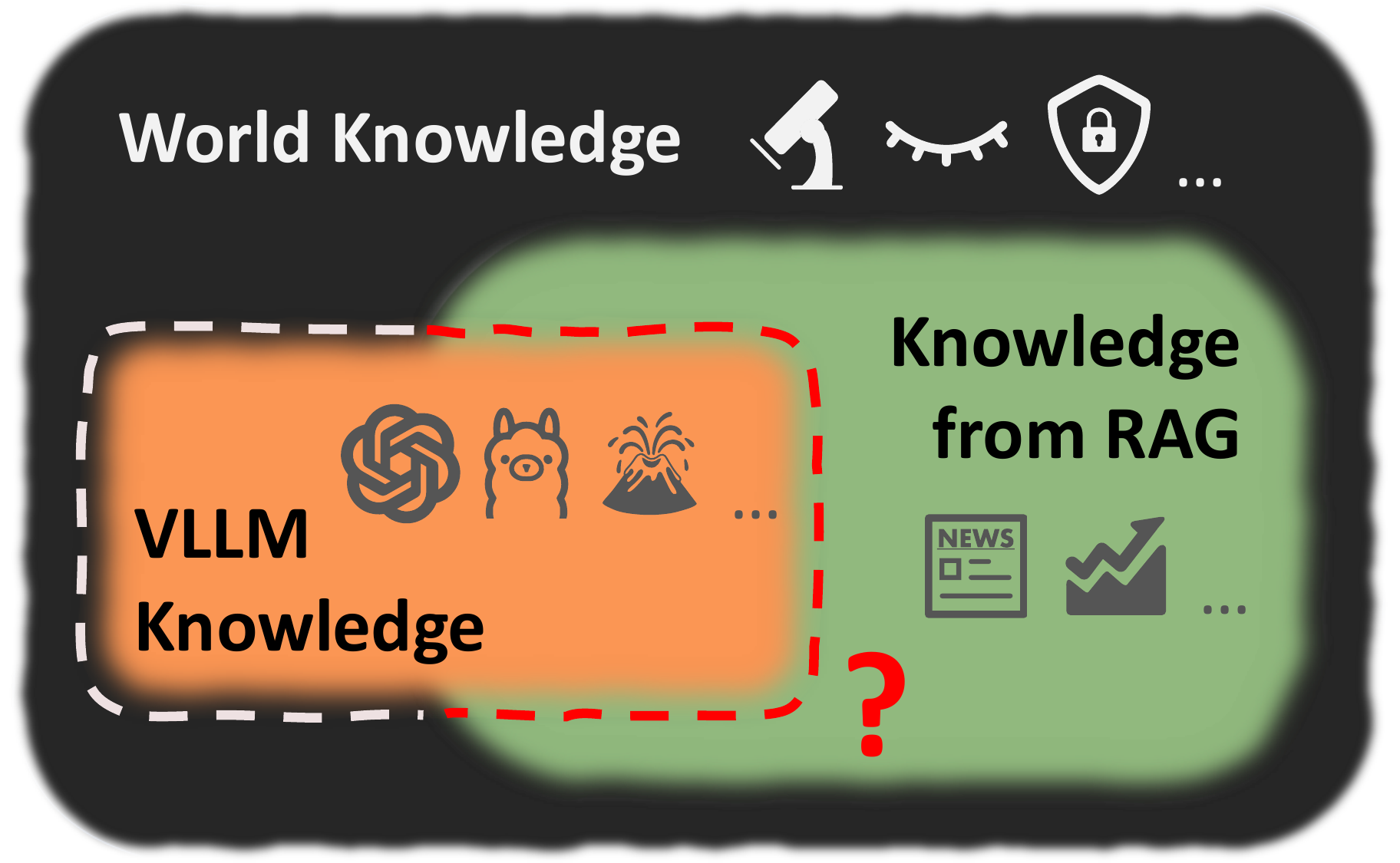}
\caption{VLLMs Knowledge Boundary concept. The black part represents all the knowledge humans have explored, and the orange and green parts represent knowledge possessed by VLLMs and knowledge that can be retrieved from external sources respectively. They overlap in some areas and the boundary between them remains unclear. The overall knowledge boundary of VLLMs can be differentiated into two parts that overlap with knowledge between RAG and world knowledge. Our method aims to identify both, and we conduct experiments to validate the potential VQA performance improvements using RAG.}
\label{outline}
\end{figure}

Some works study the knowledge boundary of text LLMs \cite{li2025refine, cheng2024can, zhang2024exploring, ren2023investigating} via prompt-based or SFT-based methods.
As of yet, there has been little research on the methodology for determining the knowledge boundaries of VLLMs. In practical applications, to answer VQA queries outside its knowledge scope, indiscriminately employing Retrieval Augmented Generation (RAG) techniques is often a viable solution. 
Although this approach has been proven to enhance the (V)LLMs' performance \cite{wang2021improving, lewis2020retrieval, chen-etal-2017-reading}, the comprehensive reliance on retrieval methods incurs significant latency due to the retrieval steps and the introduction of excessively long inputs \cite{chevalier2023adapting, zhang2024longcontextcompressionactivation, chen-etal-2024-improving-retrieval}.

To mitigate the dependence on retrieval for answering questions and simultaneously maintain, or improve, the performance benefits provided by retrieval, we aim to develop a method that can depict the knowledge boundary of a VLLM. 
In this paper, we employ a method with two variants to delineate the knowledge boundaries of a VLLM by fine-tuning a VLLM on data constructed based on sampling the responses of the VLLM.

With the ability to depict the knowledge boundary of a VLLM, we then adopt RAG techniques to validate the accuracy of the identified boundary in various held-out datasets. 
We conduct experiments using a variety of VQA datasets, including three knowledge-intensive datasets, two non-knowledge-intensive datasets, and one mixed dataset. After determining whether a query falls within the knowledge boundary, we use RAG to assess the potential improvements the retrieved information provides to the queries falling out of the knowledge boundary. Our experimental results reveal that on a mixed dataset, which contains both non-knowledge-intensive and knowledge-intensive queries simulating real situations, our method outperforms the indiscriminative use of RAG (denoted ``All RAG'') and prompt-based baseline with 50.67\% retrieval reduction. The fine-tuned knowledge boundary model lowers the retrieving ratio on less knowledge-intensive data and obtains close or even better performance compared to the ``All RAG'' setting. 
Besides, we show that the fine-tuned VLLM for boundary identification for one VLLM can be used as a surrogate boundary identifier for other VLLMs.

To sum up, our contributions are as follows:
\begin{enumerate}[leftmargin=*,noitemsep, topsep=1pt]
    \item We propose a method with two variants that detects the knowledge boundary of a VLLM.
    \item Experimental results show that we maintain, or even improve, the performance of the VLLM on various types of data while lowering the ratio of using RAG, and our method outperforms the ``All RAG'' setting and other baselines on a dataset simulating real situations.
    \item We show that the knowledge boundary for one VLLM can be used as a surrogate boundary for other VLLMs, to reduce retrieval while maintaining or improving the performance.
\end{enumerate}

\section{Method}
\label{method}

\begin{figure*}[tb]
\centering
\includegraphics[width=0.8\linewidth]{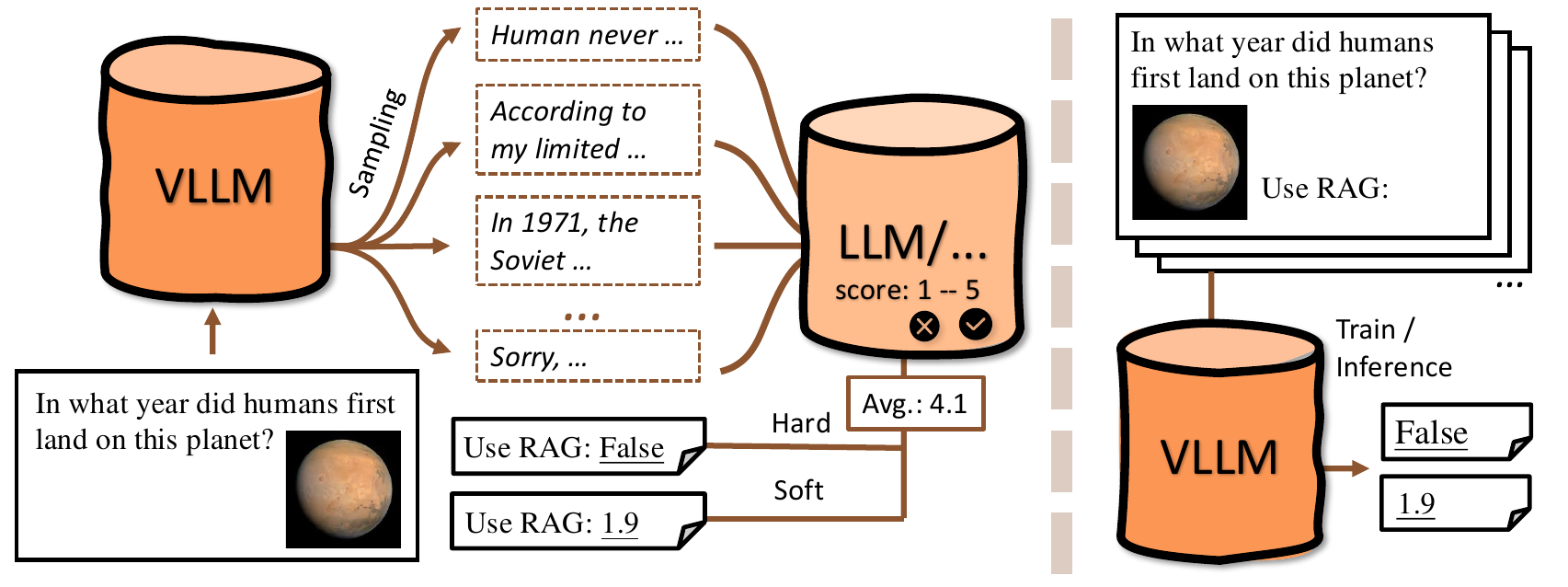}
\caption{Method illustration of training a Knowledge Boundary model.}
\label{method_fig}
\end{figure*}

We propose a method with two variants that fine-tunes a VLLM, which can depict the \textit{hard} or \textit{soft} knowledge boundary of VLLMs. The proposed method relies only on (V)LLMs and does not require manual annotation. 
In the following sections, we first introduce the background and necessary notations. Then we give details on constructing two types of datasets for fine-tuning a VLLM for knowledge boundary approximation.

\subsection{Background}
Consider a Visual Question Answering query $\bm{q}$ with gold text answer $\bm{a}$, where $\bm{q}$ contains image(s) $\bm{q_i}$ and a text query  $\bm{q_t}$. Also, contexts $\bm{k}$ related to $\bm{q}$ can be retrieved from a given corpus, where $\bm{k}$ can refer to the collection of both texts and images. Given a VL model, parameterized by $\theta$, we can answer the query with or without RAG by running decoding $(Dec)$ on the model:
\begin{equation}
\label{norag_all_rag}
    \begin{aligned}
    \bm{y_n} &= Dec_\theta(\bm{y}|\bm{q}) \\
    \bm{y_r} &= Dec_\theta(\bm{y}|\bm{q}, \bm{k})
    \end{aligned}
\end{equation}
where $\bm{k}$ might also contain prompts connecting related content and it is omitted here for simplicity.

It is acknowledged that VLLMs have a limited knowledge scope \cite{lin-byrne-2022-retrieval, wu2022multi}, denoted as $S$, and the boundary is a rather vague concept and is hard to depict accurately. 

\subsection{Sampling}
\label{sampling_section}

To approximate whether a query $\bm{q}$ should lie in VLLMs' knowledge scope $S$, we run repeated sampling of a VLLM and collect its outputs. The sampling methods include but are not limited to, top-p sampling and top-k sampling. These sampling-based methods are widely adopted to study the model's knowledge boundary problems \cite{li2025refine, zhang2024exploring, cheng2024can}. By running $R$ times sampling, we obtain $R$ outputs given query $\bm{q}$:
\begin{align}\label{sampling}
\bm{y^{(i)}} &= Dec_\theta(\bm{y}|\bm{q}), i \in \{1, 2, ..., R\}
\end{align}
After obtaining the $R$ predictions, a text LLM is prompted\footnote{The prompt is referenced from \citet{Liu_LlamaIndex_2022}. Please refer to our code for a detailed definition.} to evaluate each prediction $y^{(i)}$ where the gold answer is also given. Subsequently a score $s_i \in [s_w, s_c]$ is provided by this text LLM. We define the score range within $s_w$ and $s_c$, where $s_c$ indicates a perfectly correct answer and $s_w$ indicates a wrong answer. Then an average score is calculated over $R$ scores, indicating the overall performance of this query:
\begin{align}\label{score}
s &= mean(s_i), i \in \{1, 2, ..., R\}
\end{align}
and we note that $s$ is also $\in [s_w, s_c]$. 

\subsection{Training}
\label{training_section}

The score $s$ is used to construct the knowledge boundary training data. We differentiate our method into two variants. 
A VLLM is adopted to train on the knowledge boundary training data. We denote the parameters by $\phi$.

\paragraph{Hard Knowledge Boundary} By setting a threshold $\epsilon$, we deem the queries with score $s \geq \epsilon$ inside the knowledge boundary $S$ and the rest outside $S$. The query $\bm{q}$, together with proper prompts $P_h$, will be constructed into a training sample $\bm{x}(\bm{q}, P_h)$ as shown in Sec.~\ref{training_example_appendix_section}. For any $\bm{x}(\bm{q}, P_h)$ in the training dataset, we define the training objective $J_h$ w.r.t. $\phi$ as follows: 
\begin{equation}
\label{objective_hard}
    \begin{aligned}
    J_h (\phi) &= - \sum\limits_{\bm{x}(\bm{q}, P_h): \bm{q} \notin S} \log P_\phi(\text{``True''}|\bm{x}(\bm{q}, P_h)) \\
    &- \sum\limits_{\bm{x}(\bm{q}, P_h): \bm{q} \in S} \log P_\phi(\text{``False''}|\bm{x}(\bm{q}, P_h))
    \end{aligned}
\end{equation}
where $P_\phi(a|b)$ stands for the probability model $\phi$ predicts on $a$ given input $b$. $\phi$ is optimized by minimizing $J_h(\phi)$.

\paragraph{Soft Knowledge Boundary}
Setting a threshold to binarily classify the queries might be an overly rigid method and there is no room for adjustment when the knowledge boundary model performs poorly in possibly unseen scenarios unless we adjust $\epsilon$ and retrain the model. Thus, we also propose a method that can depict a softer boundary. Recall that for query $\bm{q}$, the average score $s$ over $R$ model predictions ranges in $[s_w, s_c]$, where $s_w$ indicates a wrong answer and $s_c$ indicates a correct one. We linearly flip the score, for example, the new score $s^{'}=s_w$ represents a strong tendency for external knowledge while $s^{'}=s_c$ represents a refusal to external knowledge.

The query $\bm{q}$, together with prompts $P_s$, will be constructed into a training sample $\bm{x}(\bm{q}, P_s)$ as shown in Sec.~\ref{training_example_appendix_section}. For any $\bm{x}(\bm{q}, P_s)$ in the training dataset, we define the training objective as follows:
\begin{align}\label{objective_soft}
J_s(\phi) = - \sum\limits_{\bm{x}(\bm{q}, P_s)} \log P_\phi(s^{'}|\bm{x}(\bm{q}, P_s))
\end{align}
where $\phi$ is optimized by minimizing $J_s(\phi)$.

By optimizing objective \ref{objective_hard}, we get a Hard Knowledge Boundary model $HKB_{\phi}$ that can take a VQA sample and predict a \textit{binary} output ``True'' or ``False'' indicating whether the RAG technique can help solve this query. Similarly, a Soft Knowledge Boundary model $SKB_{\phi}$ that can predict a \textit{soft score}, ranging from $s_w$ to $s_c$, is trained by optimizing objective \ref{objective_soft}:
\begin{equation}
    \begin{aligned}\label{KB}
    HKB_{\phi}(\bm{x}(\bm{q},P_h)) &= \text{True / False} \\
    SKB_{\phi}(\bm{x}(\bm{q},P_s)) &\in [s_w, s_c]
    \end{aligned}
\end{equation}

\subsection{Application of RAG in Our Method}

An indicator function is defined to map the prediction of a Hard/Soft Knowledge Boundary model to a real search decision:
\begin{equation}
\label{search_decision}
    \mathbb{I}(\bm{q}, \bm{k})=\left\{
    \begin{aligned}
    & \bm{k}, \text{if } HKB_{\phi}(\bm{x}(\bm{q},P_h)) == \text{true} \\
    & \quad \text{ or } SKB_{\phi}(\bm{x}(\bm{q},P_s)) \geq \epsilon\\
    & None, \text{else}
    \end{aligned}
    \right.
\end{equation}

Then we can combine the decoding with or without RAG stated in equation \ref{norag_all_rag} into:
\begin{equation}
    \begin{aligned}
    \label{KB_decoding}
    \bm{y_{kb}} &= Dec_{\theta,\phi}(\bm{y}|\bm{q}, \mathbb{I}(\bm{q}, \bm{k}))
    \end{aligned}
\end{equation}

\section{Experiment}

\subsection{Setup}

\subsubsection{Training Data}
\label{training_time_section}

With method stated in Sec.~\ref{sampling_section} and \ref{training_section}, we adopt InfoSeek \cite{chen2023can}, OK-VQA \cite{marino2019ok}, VQAv2.0 \cite{goyal2017making}, MMBench \cite{liu2025mmbench}, and MME \cite{fu2023mme} to construct the training set where we randomly sample two subsets from InfoSeek and VQAv2.0 respectively due to their large sizes. Table~\ref{training_set_table} presents the detailed sizes for each dataset we use along with the average scores $s$. In our experiments $s_w=1$ and $s_c=5$. We adopt all these datasets to increase the diversity of queries as much as possible. A detailed description of each dataset is stated in Sec.~\ref{training_data_description}.

\begin{table}[t]
\centering
\small
\scalebox{1}{
\begin{tabular}{lccc} 
\toprule
\textbf{Source} & \textbf{\# Samples} & \textbf{Model} & \textbf{Avg. Score $\pm$ std.} \\ 
\midrule
\multirow{2}{*}{InfoSeek} & \multirow{2}{*}{$216000$} & QW & $1.82 \space \pm$ \scriptsize{$1.17$} \\
 &  & DS & $1.86 \space \pm$ \scriptsize{$1.28$} \\ 
\midrule
\multirow{2}{*}{OK-VQA} & \multirow{2}{*}{$9009$} & QW & $3.70 \space \pm$ \scriptsize{$1.48$} \\
 &  & DS & $4.92 \space \pm$ \scriptsize{$0.47$} \\ 
\midrule
\multirow{2}{*}{VQAv2.0} & \multirow{2}{*}{$108000$} & QW & $4.27 \space \pm$ \scriptsize{$1.36$} \\
 &  & DS & $4.50 \space \pm$ \scriptsize{$1.22$} \\ 
\midrule
\multirow{2}{*}{\begin{tabular}[c]{@{}l@{}}MMBench \\(en)\end{tabular}} & \multirow{2}{*}{$4329$} & QW & $3.92 \space \pm$ \scriptsize{$1.72$} \\
 &  & DS & $4.08 \space \pm$ \scriptsize{$1.65$} \\ 
\midrule
\multirow{2}{*}{MME} & \multirow{2}{*}{$2374$} & QW & $4.15 \space \pm$ \scriptsize{$1.63$} \\
 &  & DS & $4.15 \space \pm$ \scriptsize{$1.64$} \\
\bottomrule
\end{tabular}}
\caption{Training set sources and statistics. Answers are sampled from Qwen-VL-7B-Chat (QW) \cite{Qwen-VL} DeepSeek-VL-7B-Chat (DS) \cite{lu2024deepseekvl} respectively. Scores are evaluated by Qwen-Max \cite{qwen1.5}}
\label{training_set_table}
\end{table}

\subsubsection{Test Data}

As we aim to construct a model that can take various input queries and make good judgments about the knowledge boundary, we adopt held-out data to evaluate the final VQA performance. We summarize the overall RAG Effect on each data in Table~\ref{test_set_table} and a brief introduction as follows.

\paragraph{Life VQA} We collect a set of VQA data from people's daily lives and extract the ones current VLLMs do not perform well, which is used to verify whether our model decides to resort to RAG for help. We will release this data along with the code and name it Life VQA. 

\paragraph{Private VQA} is an internal dataset spanning broad categories, including animals, plants, architecture, geographic locations, etc. Due to the complexity of the backgrounds and the presence of multiple objects, this collection poses a notable challenge for advanced visual reasoning and understanding. This dataset will not be released for now.

\paragraph{Dyn-VQA} is released by \citet{li2024benchmarkingmultimodalretrievalaugmented} and contains three types of questions: questions with rapidly changing answers, questions requiring multi-modal knowledge and multi-hop questions. This dataset is a challenging one in our evaluation. \textit{Gold query} is annotated by \citet{li2024benchmarkingmultimodalretrievalaugmented} that combines the text query and image to be used to retrieve useful information.

\paragraph{NoCaps} \cite{agrawal2019nocaps} is an open-domain image captioning dataset derived from Open Images \cite{openimages}, focusing on generating captions for a diverse array of objects and scenes. We sample a subset of size 500. 

\paragraph{Visual7W} \cite{zhu2016visual7w} is a VQA dataset containing images from COCO \cite{lin2014microsoft}, paired with seven types of questions (who, what, when, where, how, why and which). It aims to evaluate models' abilities in object recognition and deeper reasoning within visual contexts.

\paragraph{Mix} is a composite dataset consisting of 100 samples from each of the aforementioned datasets. It is designed to integrate the characteristics of each dataset and simulate real-world scenarios. Thus the effect of RAG on this dataset is mixed and hard to predict intuitively.


\begin{table}[t]
\centering
\small
\scalebox{1.0}{
\begin{tabular}{lc} 
\toprule
\textbf{Test Data} & \textbf{RAG Effect} \\
\midrule
Life VQA & High   \\ 
\midrule
Private VQA & Medium  \\ 
\midrule
Dyn-VQA & High \\
\midrule
NoCaps & Low \\
\midrule
Visual7W & Low \\
\midrule 
Mix & ? \\
\bottomrule
\end{tabular}}
\caption{Test data property illustration of whether RAG is helpful in answering the queries.}
\label{test_set_table}
\end{table}

\subsubsection{Use of RAG} We aim not only to locate the queries that need RAG to answer better but also to adopt retrieval techniques to verify the final VQA performance with the search decision $HKB_\phi$ and $SKB_\phi$ defined in equations \ref{KB}. We note that although there are various options for retrieval, such as text search and image search, we do not design detailed methods to determine the best option in this paper. Instead, we directly use text search (Google) for Dyn-VQA and image search (Bing) for the rest for better retrieval information quality towards answering the question. We note that Dyn-VQA is a challenging dataset that exhibits multi-hop property, therefore we use the \textit{golden query} \citet{li2024benchmarkingmultimodalretrievalaugmented} have summarized for retrieving useful information.

In the following sections, the ``No RAG'' setting refers to the performance of only VLLMs and no retrieval information is given, and ``All RAG'' refers to always incorporating RAG. ``Prompt-based'' refers to prompting the model that is sampled to adopt RAG or not.

\begin{table*}
\centering
\scalebox{0.8}{
\begin{tabular}{llcccccccc|cc} 
\toprule
\multicolumn{1}{l}{\textbf{Dataset}} & \textbf{Metric} & \begin{tabular}[c]{@{}c@{}}\textbf{No }\\\textbf{RAG}\end{tabular} & \begin{tabular}[c]{@{}c@{}}\textbf{All }\\\textbf{RAG}\end{tabular} & \begin{tabular}[c]{@{}c@{}}\textbf{Prompt-}\\\textbf{based}\end{tabular} & \textbf{\%} & \textbf{HKB} & \textbf{\%} & \textbf{SKB} & \textbf{\%} & \textbf{Human} & \textbf{\%} \\ 
\midrule
\multirow{2}{*}{\textbf{Life VQA}} & \textbf{LLM} & 30.00 & 40.70 & 33.89 & 12.75\% & 40.64 & 96.64\% & 36.78 & 61.74\% & 39.33 & 71.14\% \\
  & \textbf{Acc.} & 17.80 & 36.11 & 21.38 & 12.75\% & 36.11 & 96.64\% & 29.44 & 61.74\% & 33.36 & 71.14\% \\
\midrule
\multirow{2}{*}{\textbf{Private VQA}} & \textbf{LLM} & 22.90 & 24.35 & 24.95 & 14.80\% & 24.50 & 99.20\% & 22.89 & 67.80\% & 24.20 & 72.00\% \\
  & \textbf{Acc.} & 16.26 & 18.40 & 17.26 & 14.80\% & 18.40 & 99.20\% & 17.35 & 67.80\% & 18.55 & 72.00\% \\
\midrule
\multirow{2}{*}{\textbf{Dyn-VQA ch}} & \textbf{LLM} & 19.16 & 38.95 & 19.70 & 6.38\% & 37.94 & 95.66\% & 36.53 & 84.26\% & 28.89 & 46.95\% \\
  & \textbf{Acc.} & 23.41 & 43.06 & 24.37 & 6.38\% & 42.71 & 95.66\% & 40.97 & 84.26\% & 33.13 & 46.95\% \\
\midrule
\multirow{2}{*}{\textbf{Dyn-VQA en}} & \textbf{LLM} & 21.60 & 34.93 & 23.51 & 14.13\% & 33.30 & 89.79\% & 32.06 & 76.08\% & 25.73 & 29.51\% \\
  & \textbf{Acc.} & 25.64 & 41.87 & 27.58 & 14.13\% & 40.66 & 89.79\% & 38.51 & 76.08\% & 30.83 & 29.51\% \\
\midrule
\multirow{2}{*}{\textbf{NoCaps}} & \textbf{LLM} & 50.13 & 30.37 & 50.13 & 0.00\% & 42.50 & 38.40\% & 50.13 & 0.00\% & 50.13 & 0.00\% \\
  & \textbf{Acc.} & 40.50 & 30.72 & 40.50 & 0.00\% & 36.95 & 38.40\% & 40.50 & 0.00\% & 40.50 & 0.00\% \\
\midrule
\multirow{2}{*}{\textbf{Visual7W}} & \textbf{LLM} & 54.48 & 52.04 & 55.32 & 31.36\% & 52.95 & 35.37\% & 54.27 & 2.96\% & 54.53 & 0.52\% \\
  & \textbf{Acc.} & 44.34 & 44.94 & 44.18 & 31.36\% & 44.32 & 35.37\% & 44.68 & 2.96\% & 44.34 & 0.52\% \\
\midrule
\midrule
\multirow{2}{*}{\textbf{Mix}} & \textbf{LLM} & 34.44 & 38.60 & 34.98 & 12.67\% & \uline{39.59} & 76.83\% & \textbf{39.93} & 49.33\% & 38.29 & 38.33\% \\
  & \textbf{Acc.} & 26.13 & 32.39 & 27.23 & 12.67\% & \textbf{32.73} & 76.83\% & 30.98 & 49.33\% & 31.02 & 38.33\% \\
\bottomrule
\end{tabular}
}
\caption{Main results of Qwen-VL-Chat. Scores are shown in columns except for the \% ones. Metrics are evaluated by Qwen-Max (LLM) and Token Accuracy (Acc.). \underline{Underlines} mark the results that outperform three baseline ``No RAG'', ``All RAG'' and ``Prompt-based'' settings. \textbf{Boldface} marks the best results.}
\label{main_results_7b_table}
\end{table*}

\subsubsection{Base Models}
When constructing the training set according to the method stated in Sec.~\ref{sampling_section}, we experiment with Qwen-VL-7B-Chat and DeepSeek-VL-7B-Chat that are used to be sampled $R=30$ times and fine-tuned according to Sec.~\ref{training_section} respectively. Refer to Sec.~\ref{hyper_appendix_section} for detailed training settings. Qwen-Max is prompted to score the $R$ predictions to get scores $s_i$ where we adopt $s_w=1$ and $s_c = 5$ referenced from \citet{Liu_LlamaIndex_2022}.

For Visual Question Answering, we first evaluate the performance of the original models to be sampled. In addition, we seek to validate whether the identified knowledge boundary can function as a surrogate boundary for other VLLMs since constructing training datasets through sampling (Sec.~\ref{training_section}) on (larger) models can be prohibitively expensive. We further validate the surrogate knowledge boundary on the following VLLMs, Qwen-VL-Max \cite{Qwen-VL}, Qwen-VL-2 \cite{Qwen2-VL} and GPT-4o \cite{hurst2024gpt}, to evaluate its potential for generalizing across different VLLMs.

\subsection{Main Results}
\label{main_result}

We present our main results of Qwen-VL-7B-Chat in Table~\ref{main_results_7b_table} and results of DeepSeek-VL-7B-Chat in Appendix~\ref{main_results_ds_section}. In this section, we focus on the results of Qwen.

Metrics \textbf{LLM} represents that the score is evaluated by a text LLM, Qwen-Max, given the model prediction and gold answer. Metrics \textbf{Acc.} refers to token accuracy which involves determining the proportion of tokens in the model's predictions that match the tokens in the gold answer. Both Scores range from 0 to 100 and a higher score indicates a higher performance. The \% columns refer to the ratio of data that our knowledge boundary model predicts to lie beyond the VLLM's knowledge boundaries. The ``Human'' column represents the corresponding statistics where the Knowledge Boundary model is trained on the human-labeled data mentioned in Sec.~\ref{training_time_section}, and we deem it a reference result. 

First, the results in the Mix row, which considers all kinds of VQA queries in our setting and simulates a real situation, show that our methods outperform all other baseline and reference settings. Our $HKB$ method lowers the retrieval demand by 23.17\%, and the $SKB$ method lowers it by 50.67\%.

Second, as shown by the \% columns and the RAG Effect we summarized in Table~\ref{test_set_table}, our Knowledge Boundary models succeed in predicting a high ratio on test data when RAG can effectively aid in answering the query, and it lowers the ratio for data where the queries tend to fall within the knowledge scope of a VLLM.

\begin{table*}[t]
    \centering
    \scalebox{0.75}{
    \begin{tabular}{llcccccccc|cc} 
    \toprule
    \multicolumn{1}{c}{} & \multicolumn{1}{c}{\begin{tabular}[c]{@{}c@{}}\textbf{Metric:}\\\textbf{LLM}\end{tabular}} & \begin{tabular}[c]{@{}c@{}}\textbf{No}\\\textbf{RAG}\end{tabular} & \begin{tabular}[c]{@{}c@{}}\textbf{All }\\\textbf{RAG}\end{tabular} & \begin{tabular}[c]{@{}c@{}}\textbf{Prompt-}\\\textbf{based}\end{tabular} & \textbf{\%} & \textbf{HKB} & \textbf{\%} & \textbf{SKB} & \textbf{\%} & \textbf{Human} & \textbf{\%} \\ 
    \midrule
    \multirow{4}{*}{\textbf{Life VQA}} & Ds.-VL-Chat & 25.54 & 47.38 & 27.68 & 12.75\% & 46.91 & 96.64\% & 41.21 & 61.74\% & 41.61 & 71.14\% \\
     & Qwen-VL-Max & 43.26 & 56.38 & 45.97 & 12.75\% & 56.85 & 96.64\% & 53.86 & 61.74\% & 55.23 & 71.14\% \\
     & Qwen-VL-2 & 42.55 & 54.43 & 46.28 & 12.75\% & 54.03 & 96.64\% & 52.28 & 61.74\% & 53.96 & 71.14\% \\
     & GPT-4o & 47.52 & 55.47 & 48.26 & 12.75\% & 56.14 & 96.64\% & 54.83 & 61.74\% & 54.90 & 71.14\% \\ 
    \midrule
    \multirow{4}{*}{\textbf{Private VQA}} & Ds.-VL-Chat & 23.01 & 27.06 & 23.89 & 14.80\% & 26.94 & 99.20\% & 26.19 & 67.80\% & 25.83 & 72.00\% \\
     & Qwen-VL-Max & 35.20 & 41.90 & 38.30 & 14.80\% & 41.68 & 99.20\% & 40.45 & 67.80\% & 43.18 & 72.00\% \\
     & Qwen-VL-2 & 35.16 & 38.02 & 36.57 & 14.80\% & 37.84 & 99.20\% & 35.85 & 67.80\% & 38.25 & 72.00\% \\
     & GPT-4o & 39.70 & 38.21 & 40.06 & 14.80\% & 37.85 & 99.20\% & 38.83 & 67.80\% & 40.21 & 72.00\% \\ 
    \midrule
    \multirow{4}{*}{\textbf{Dyn-VQA ch}} & Ds.-VL-Chat & 21.62 & 44.10 & 22.98 & 6.38\% & 42.92 & 95.66\% & 40.99 & 84.26\% & 34.24 & 46.95\% \\
     & Qwen-VL-Max & 32.97 & 51.24 & 34.23 & 6.38\% & 50.86 & 95.66\% & 48.24 & 84.26\% & 43.33 & 46.95\% \\
     & Qwen-VL-2 & 32.78 & 50.74 & 34.02 & 6.38\% & 50.48 & 95.66\% & 48.19 & 84.26\% & 43.05 & 46.95\% \\
     & GPT-4o & 41.91 & 56.31 & 42.53 & 6.38\% & 56.31 & 95.66\% & 54.49 & 84.26\% & 48.95 & 46.95\% \\ 
    \midrule
    \multirow{4}{*}{\textbf{Dyn-VQA en}} & Ds.-VL-Chat & 25.58 & 38.10 & 27.19 & 14.13\% & 36.86 & 89.79\% & 36.32 & 76.08\% & 29.44 & 29.51\% \\
     & Qwen-VL-Max & 37.19 & 43.98 & 38.32 & 14.13\% & 43.09 & 89.79\% & 42.78 & 76.08\% & 39.48 & 29.51\% \\
     & Qwen-VL-2 & 37.12 & 44.20 & 37.17 & 14.13\% & 42.47 & 89.79\% & 42.32 & 76.08\% & 40.07 & 29.51\% \\
     & GPT-4o & 45.41 & 50.93 & 45.24 & 14.13\% & 49.88 & 89.79\% & 48.75 & 76.08\% & 47.14 & 29.51\% \\ 
    \midrule
    \multirow{4}{*}{\textbf{NoCaps}} & Ds.-VL-Chat & 63.67 & 59.81 & 63.67 & 0.00\% & 61.23 & 38.40\% & 63.67 & 0.00\% & 63.67 & 0.00\% \\
     & Qwen-VL-Max & 62.10 & 49.66 & 62.10 & 0.00\% & 57.09 & 38.40\% & 62.10 & 0.00\% & 62.10 & 0.00\% \\
     & Qwen-VL-2 & 62.10 & 49.93 & 62.10 & 0.00\% & 56.93 & 38.40\% & 62.10 & 0.00\% & 62.10 & 0.00\% \\
     & GPT-4o & 61.43 & 63.98 & 61.43 & 0.00\% & 62.12 & 38.40\% & 61.43 & 0.00\% & 61.43 & 0.00\% \\ 
    \midrule
    \multirow{4}{*}{\textbf{Visual7W}} & Ds.-VL-Chat & 58.34 & 57.29 & 57.26 & 31.36\% & 57.85 & 35.37\% & 58.13 & 2.96\% & 58.28 & 0.52\% \\
     & Qwen-VL-Max & 58.37 & 55.51 & 62.11 & 31.36\% & 57.10 & 35.37\% & 58.25 & 2.96\% & 58.30 & 0.52\% \\
     & Qwen-VL-2 & 58.16 & 54.41 & 62.19 & 31.36\% & 56.66 & 35.37\% & 57.85 & 2.96\% & 58.02 & 0.52\% \\
     & GPT-4o & 52.96 & 47.06 & 51.82 & 31.36\% & 50.87 & 35.37\% & 52.89 & 2.96\% & 52.87 & 0.52\% \\ 
    \midrule
    \midrule
    \multirow{4}{*}{\textbf{Mix}} & Ds.-VL-Chat & 34.96 & 45.18 & 35.71 & 12.67\% & 45.08 & 76.83\% & 43.35 & 49.33\% & 42.20 & 38.33\% \\
     & Qwen-VL-Max & 46.54 & 49.26 & 47.30 & 12.67\% & \uline{50.64} & 76.83\% & \uline{51.06} & 49.33\% & 52.05 & 38.33\% \\
     & Qwen-VL-2 & 46.36 & 47.89 & 47.46 & 12.67\% & \uline{49.31} & 76.83\% & \uline{49.29} & 49.33\% & 51.41 & 38.33\% \\
     & GPT-4o & 51.44 & 52.90 & 50.57 & 12.67\% & \uline{54.10} & 76.83\% & \uline{52.97} & 49.33\% & 55.27 & 38.33\% \\
    \bottomrule
    \end{tabular}
    }
    \caption{Knowledge Boundary model (Qwen-VL-7B-Chat) as a surrogate boundary identifier for other VLLMs. }
    \label{main_results_llm_table}
\end{table*}

Third, on the first four datasets where RAG can (greatly) enhance the VQA performance, we show that with our $HKB$ and $SKB$, the performance is close to that achieved with the ``All RAG'' setting. For example, with the $SKB$ model, Qwen-VL-Chat archives a 32.06 LLM score on the Dyn-VQA (en) dataset with 76.08\% RAG ratio, whereas the ``All RAG'' setting achieves 34.93. With the $HKB$ model, Qwen-VL-Chat exceeds the ``All RAG'' setting on Private VQA, even though we note that ``All RAG'' is a strong setting on this data. 

At last, on the NoCaps and Visual7W datasets where VLLMs can perform well without RAG and RAG tends to supply noise, our method can identify a much lower search ratio. Specifically, the search ratio from $SKB$ is close to or equal to zero.

\section{Analysis}
\label{analysis}

In this section, we present five analytical experiments. 
The first shows the performance of other VLLMs if we employ the identified knowledge boundary as a surrogate. The second shows how the RAG ratio and VQA performance are affected by the threshold defined in the $SKB$ variant. The third presents the efficiency of our method. 
The fourth is a case study showing cases with different Knowledge Boundary model predictions (inside or outside the knowledge boundary).
The last shows the accuracy of VLLM boundary identification on held-in data at training time.

\begin{figure*}[t]
    \centering
    \includegraphics[width=0.32\linewidth]{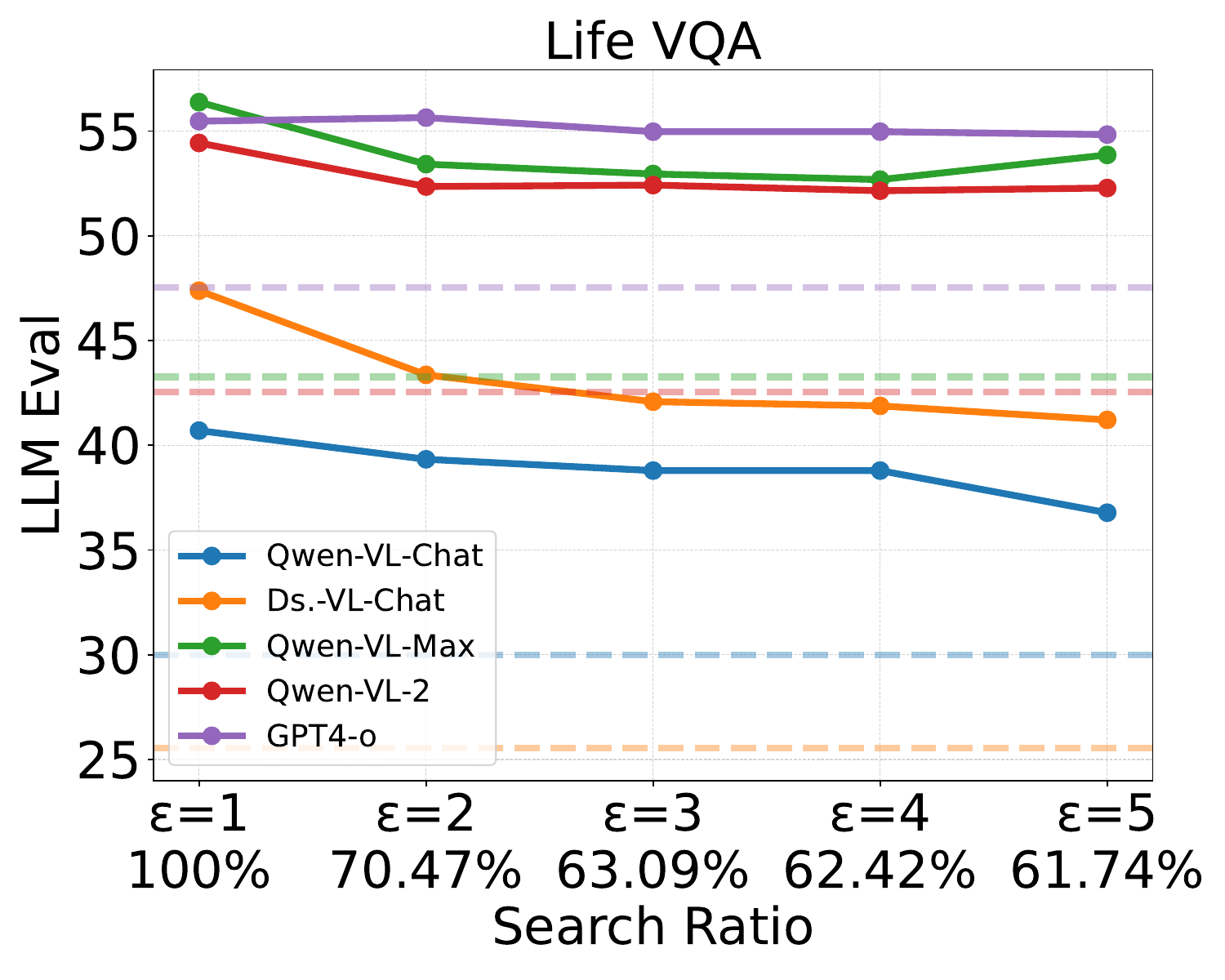}
    \includegraphics[width=0.32\linewidth]{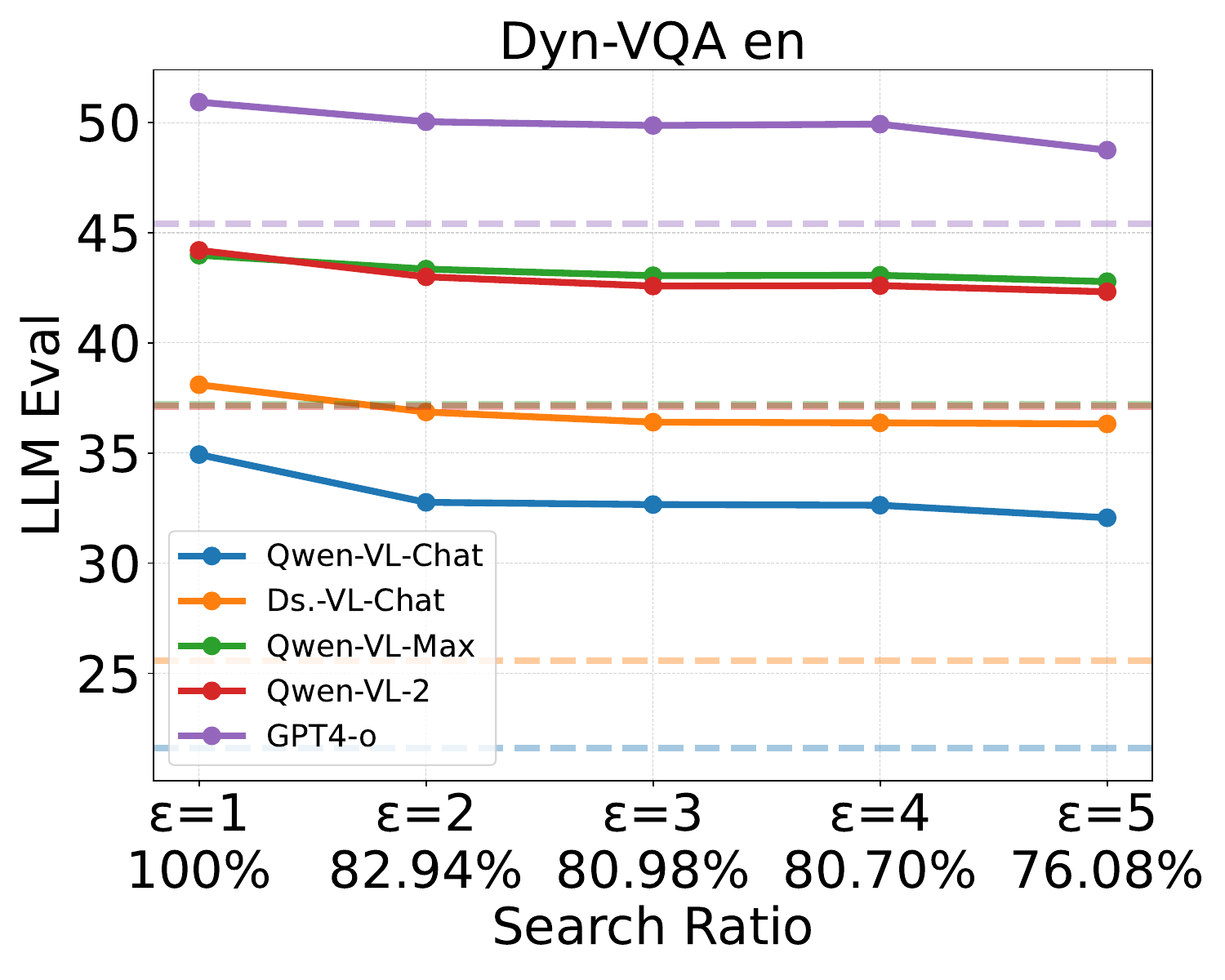}
    \includegraphics[width=0.32\linewidth]{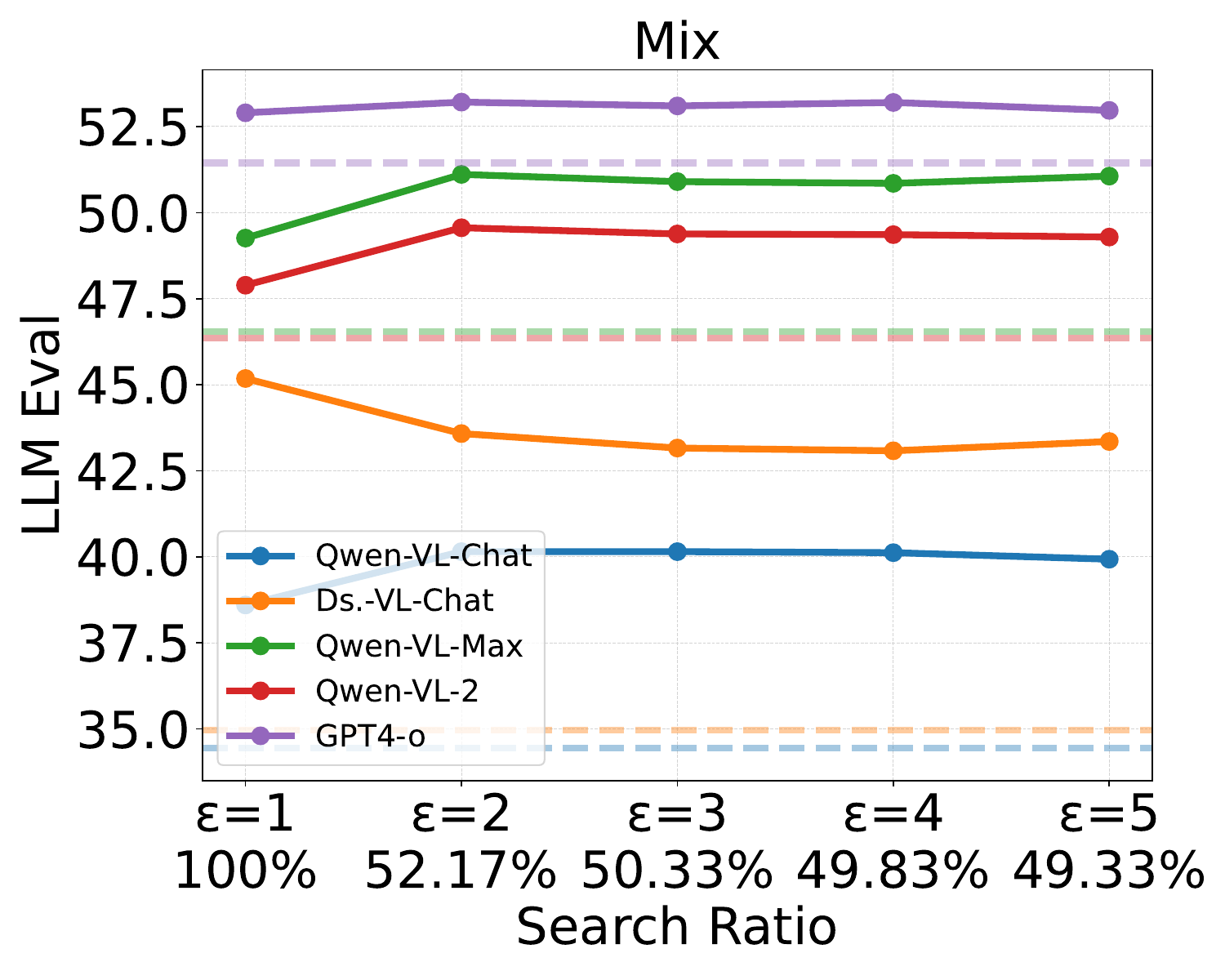}
    \includegraphics[width=0.32\linewidth]{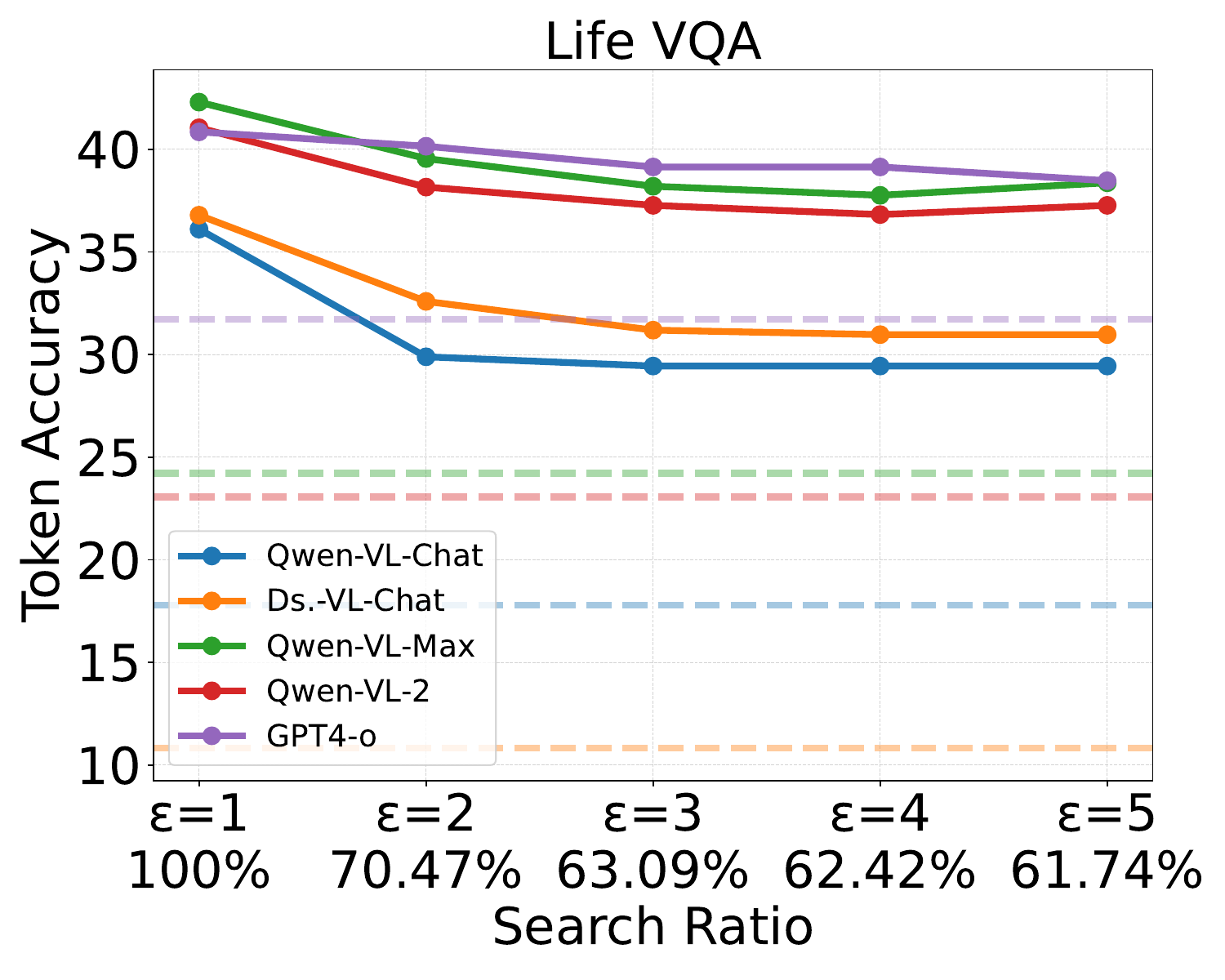}
    \includegraphics[width=0.32\linewidth]{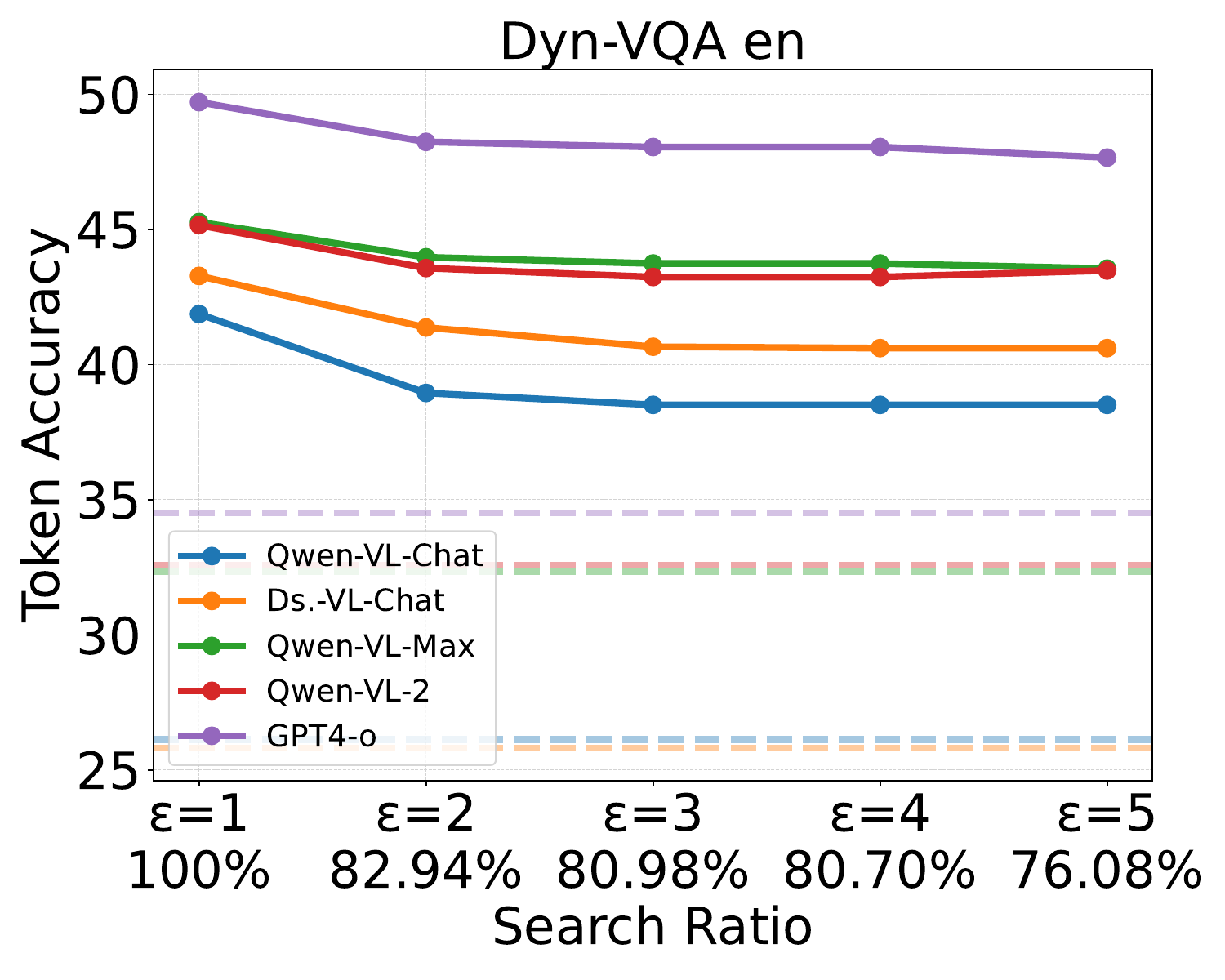}
    \includegraphics[width=0.32\linewidth]{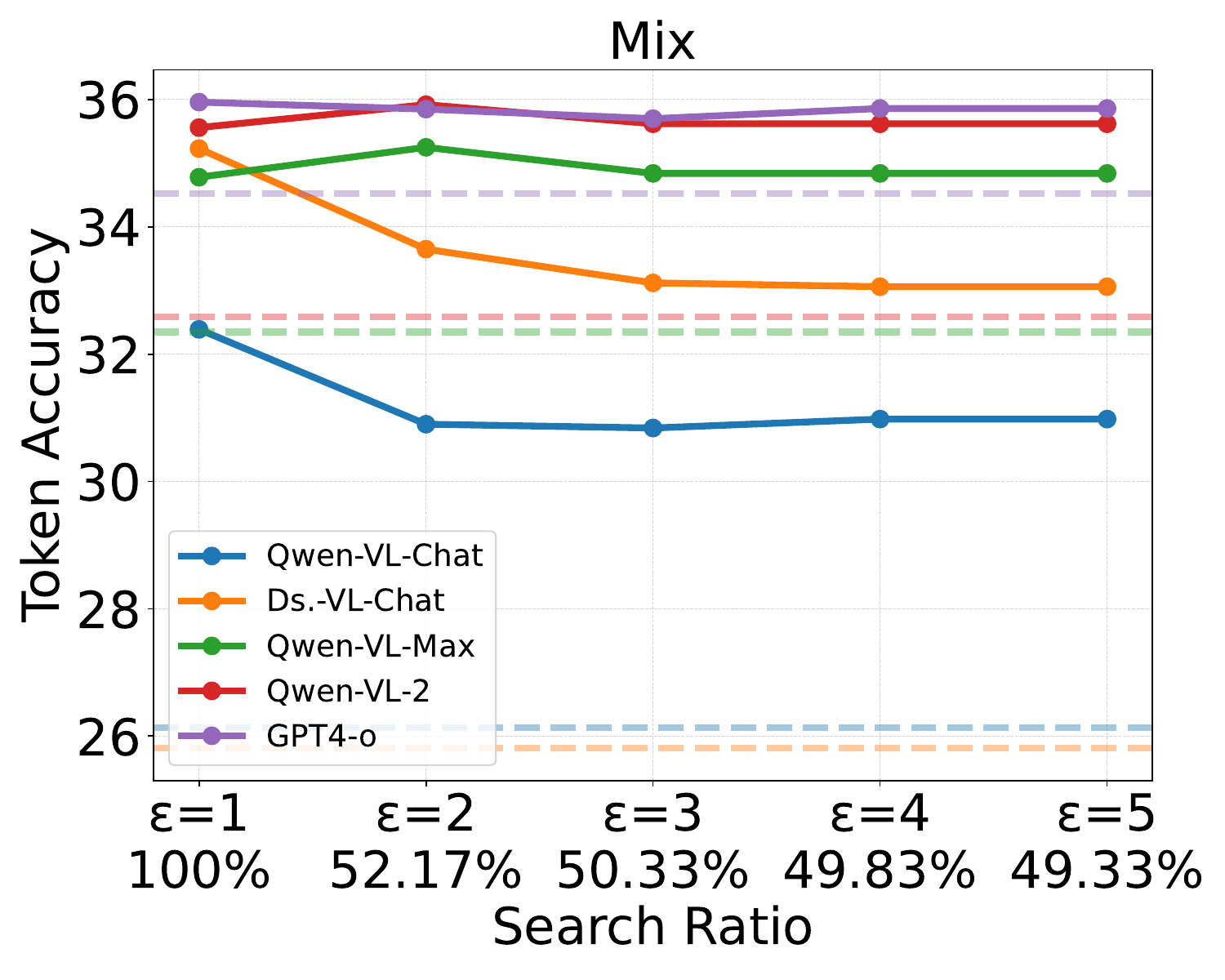}
    \caption{Effect of $\epsilon$. The lighter dashed lines accordingly indicate the performance under each base model's ``No RAG'' setting. Knowledge Boundary model is Qwen-VL-7B-Chat.}
    \label{epsilon_fig}
\end{figure*}

\subsection{Surrogate Boundary for Other VLLMs}
\label{plugin_section}

We assemble around 340 thousand VQA samples from various domains discussed in Sec.~\ref{training_time_section}. Sampling each data thirty times is prohibitively expensive for closed-source VLLMs. 
While VLLMs (e.g., Qwen-VL, DeepSeek-VL) may intuitively exhibit different knowledge scopes, this overlap is expected given their shared pretraining corpora (e.g., LAION \cite{schuhmann2022laion}, COCO \cite{lin2014microsoft}), similar visual encoder architectures (e.g., CLIP variants), and common textual knowledge from large-scale web data.
Besides, queries regarding recently occurring news events typically fall outside the knowledge boundaries of any model. Thus, we conduct an experiment that validates whether the identified knowledge boundary can function as a surrogate boundary for other VLLMs.

The experimental results with Qwen as a boundary model are presented in Table~\ref{main_results_llm_table} and Table~\ref{main_results_acc_table}. The results with DeepSeek as a boundary model are presented in Appendix~\ref{supplementary_results_of_surrogate}.

From Table~\ref{main_results_llm_table} Mix row, Qwen-VL-Max, Qwen-VL-2 and GPT-4o achieve better performance than all three baseline settings. Deepseek-VL-7B-Chat remains competitive to the ``All RAG'' setting with LLM metric and outperforms all other settings in Table~\ref{main_results_acc_table} Mix row. For other datasets, we show that the previously identified knowledge boundary can help maintain the performance with a reduced RAG ratio. For example, GPT-4o achieves 54.83 with only 61.74\% RAG ratio while the ``All RAG'' setting achieves 55.47 on the Life VQA dataset. Deepseek-VL-7B-Chat maintains its performance on the Dyn-VQA (en) dataset compared to the ``All RAG'' setting and keeps a clear margin compared to the ``No RAG'' setting with a 23.92\% retrieving deduction.

\subsection{Effect of $\epsilon$ for $SKB$}
\label{epsilon_section}

In Sec~\ref{main_result}, we show the result of the $SKB$ method with the least RAG ratio, i.e., $\epsilon$ is set to maintain a low tendency to resort to RAG. Here we show how the overall VQA performance is affected by $\epsilon$. The results of three datasets are illustrated in Fig.~\ref{epsilon_fig}. The leftmost point of the horizontal axis corresponds to the ``All RAG'' setting (with $\epsilon=s_w$), while the rightmost point represents the minimal search ratio. Light-coloured dashed lines depict the ``No RAG'' setting. 
For the left two data in Fig~\ref{epsilon_fig}, where RAG can greatly affect the performance, our methods can maintain a clear margin between the ``No RAG'' setting and obtain a relatively stable performance with a decreased search ratio. For the Mix data where all types of data are fused, our methods can still lower the search ratio while maintaining, or improving, the performance.

\definecolor{OliveGreen}{rgb}{0.0, 0.6, 0.1}
\definecolor{BrickRed}{rgb}{0.8, 0.25, 0.33}
\begin{table*}[t]
\centering
\small
\scalebox{1.02}{
\begin{tabular}{@{}p{0.45\textwidth}@{\hspace{0.06\textwidth}}p{0.45\textwidth}@{}}
\toprule
\begin{minipage}[c]{\linewidth}
    \textbf{Cases where the model predicts to be out of the knowledge boundary}
\end{minipage}
& 
\begin{minipage}[c]{\linewidth}
    \textbf{Cases where the model predicts to be in the knowledge boundary}
\end{minipage} \\
\midrule


\begin{minipage}[t]{\linewidth}
    \noindent 
    \begin{minipage}{0.7\linewidth}
        \textbf{Question:} How many World Series titles has the team won?
    \end{minipage}%
    \begin{minipage}{0.3\linewidth}
        \centering
        \includegraphics[width=0.9\linewidth]{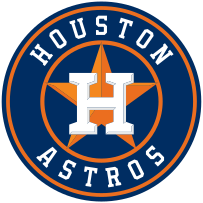} 
    \end{minipage}

    \textbf{Prediction w/o retrieval:} The Houston Astros have won one World Series title, in 2017. $\rightarrow$ (\textcolor{BrickRed}{incorrect}) \\
    \textbf{Prediction w/ retrieval:} The Houston Astros have won two World Series titles. $\rightarrow$ (\textcolor{OliveGreen}{correct})
    \vspace{0.2em}
    \textit{\footnotesize (Source: Dyn-VQA (en))}
\end{minipage}
& 
\begin{minipage}[t]{\linewidth}
    \noindent 
    \begin{minipage}{0.7\linewidth}
        \textbf{Question:} Can you add a caption to the image using a phrase? For example: A little girl in a white jacket and sandals.
    \end{minipage}%
    \begin{minipage}{0.3\linewidth}
        \centering
        \includegraphics[width=0.9\linewidth]{figs/blue_jay.jpg}
    \end{minipage}
    
    \textbf{Prediction w/o retrieval:} A blue jay perched on a tree branch. $\rightarrow$ (\textcolor{OliveGreen}{correct}) \\
    \textbf{Prediction w/ retrieval:} A blue and white bird perched on a tree branch. $\rightarrow$ (\textcolor{OliveGreen}{correct})
    \vspace{0.2em}
    \textit{\footnotesize (Source: NoCaps)}
\end{minipage} \\ 

\addlinespace[0.5em] 

\begin{minipage}[t]{\linewidth}
    \noindent 
    \begin{minipage}[c]{0.7\linewidth}
        \textbf{Question:} Who is the current Prime Minister of this country?
    \end{minipage}%
    \begin{minipage}{0.3\linewidth}
        \centering
        \includegraphics[width=0.9\linewidth]{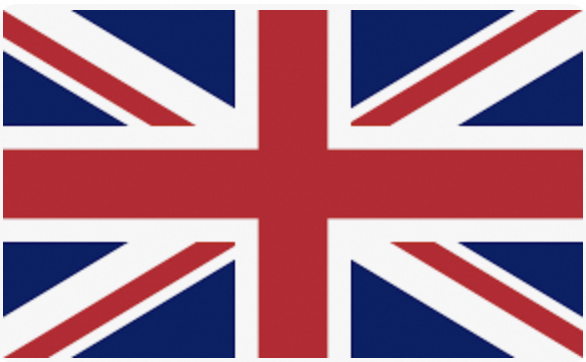}
    \end{minipage}
    
    \textbf{Prediction w/o retrieval:} The current Prime Minister of the United Kingdom is Rishi Sunak. $\rightarrow$ (\textcolor{BrickRed}{incorrect}) \\
    \textbf{Prediction w/ retrieval:} The current Prime Minister of the United Kingdom is Keir Starmer. $\rightarrow$ (\textcolor{OliveGreen}{correct})
    \vspace{0.2em}
    \textit{\footnotesize (Source: Dyn-VQA (en))}
\end{minipage}
& 
\vspace{-30pt}
\begin{minipage}[t]{\linewidth}
    \noindent 
    \begin{minipage}{0.7\linewidth}
        \textbf{Question:} How is the boat staying ashore?
    \end{minipage}%
    \begin{minipage}{0.3\linewidth}
        \centering
        \includegraphics[width=0.9\linewidth]{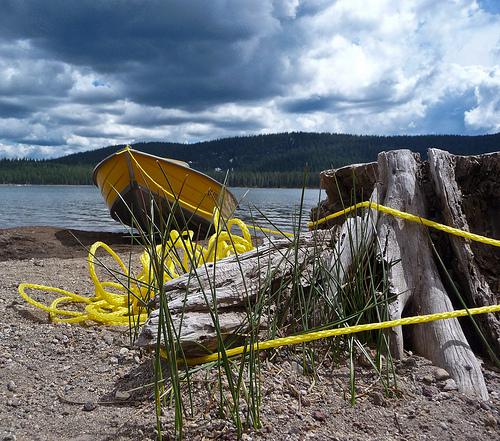}
    \end{minipage}
    
    \textbf{Prediction w/o retrieval:} The boat is staying ashore by being tied to a wooden post with a yellow rope. The rope is wrapped around the post and secured, preventing the boat from drifting back into the water. $\rightarrow$ (\textcolor{OliveGreen}{correct}) \\
    \textbf{Prediction w/ retrieval:} It is tied to a tree with rope. $\rightarrow$ (\textcolor{OliveGreen}{correct})
    \vspace{0.2em}
    \textit{\footnotesize (Source: Visual7W)}
\end{minipage} \\ 
\bottomrule
\end{tabular}
}
\caption{Cases when the model predicts whether the VQA queries are in VLLMs' knowledge boundary. Base model is Qwen-VL-Max.}
\label{tab:case_study}
\end{table*}

\subsection{Efficiency}

\begin{table}[tb]
\small
\centering
\scalebox{0.95}{
\begin{tabular}{llccc} 
\toprule
 & \textbf{Model} & \textbf{All RAG} & \begin{tabular}[c]{@{}c@{}}\textbf{HKB}\\\textbf{(Mix)}\end{tabular} & \begin{tabular}[c]{@{}c@{}}\textbf{SKB}\\\textbf{(Mix)}\end{tabular} \\ 
\midrule
\multirow{2}{*}{\textbf{Time (s) }} & \textbf{QW} & \multirow{2}{*}{619.20} & 598.13 & 427.85 \\
 & \textbf{DS} &  & 386.61 & 326.74 \\
\midrule
\multirow{2}{*}{\begin{tabular}[c]{@{}l@{}}\textbf{Improvement }\\\textbf{(\%) }\end{tabular}} & \textbf{QW} & \multirow{2}{*}{-} & 3.40\% & 30.90\% \\
 & \textbf{DS} &  & 37.56\% & 47.23\% \\
\bottomrule
\end{tabular}}
\caption{Efficiency illustration of Knowledge Boundary model Qwen-VL-7B-Chat (QW) and DeepSeek-VL-7B-Chat (DS). \textbf{Time} row shows the time spent before generating the answer in the VQA task.}
\label{efficiency_table}
\end{table}

Our method incorporates an additional forward pass for each VQA example for knowledge boundary identification. We report the overall efficiency in Table~\ref{efficiency_table} on the Mix dataset, where the All RAG setting always uses RAG (calls to the search engines included) and does not perform the forward pass, and $HKB/SKB$ refers to partially performing RAG according to our model's predictions with forward-pass time included.

\subsection{Case Study}
This section shows four cases with different Knowledge Boundary model predictions (inside or outside the knowledge boundary). In the left column of Table~\ref{tab:case_study}, the Knowledge Boundary model predicts that they are outside the knowledge boundary, and the retrieval indeed helps the model correct its response. In the right column, we show two example cases that our Knowledge Boundary model predicts to be in the knowledge boundary and thus do not need retrieval.

\subsection{Performance of Knowledge Boundary Identification on Held-In Data}

\begin{table}[t]
\centering
\small
\scalebox{1}{
\begin{tabular}{llccc} 
\toprule
\textbf{Model} & \textbf{Fold} & \textbf{Human-labeled} & \textbf{Hard} & \textbf{Soft} \\ 
\midrule
\multirow{2}{*}{\textbf{QW}} & Train & 96.25 & 90.50 & 88.41 \\
 & Val. & - & 91.16 & 88.96 \\ 
\midrule
\multirow{2}{*}{\textbf{DS}} & Train & 96.25 & 93.91 & 92.10 \\
 & Val & - & 93.76 & 92.11 \\
\bottomrule
\end{tabular}
}
\caption{Training and validation results on the held-in dataset. Metrics are shown in the accuracy defined in \texttt{ms-swift} package. We have a limited number of human-labeled samples thus we do not set a validation set for ``Human-labeled'' setting.}
\label{training_held_in_table}
\end{table}

The training results of the Knowledge Boundary model are shown in Table~\ref{training_held_in_table}. We show that by training Qwen-VL-7B-Chat (QW) and DeepSeek-VL-7B-Chat (DS), they succeed in modeling the knowledge boundary on held-in data we constructed according to Sec.~\ref{training_section}.

\section{Related Work}

\subsection{Knowledge Boundary Study of Text LLM}
As the LLMs are applied to a wider range of fields, users expect them to perform well on any query. However, inevitably, the knowledge embedded within LLMs does not automatically update over time, resulting in certain queries consistently falling outside the model’s knowledge boundaries. Some works study the Knowledge Boundaries of text LLMs. A commonly used approach prompts LLMs to output content like \textit{``I don't know''} \cite{li2025refine, cheng2024can, ren2023investigating}. Alternatively, another approach is to construct a dataset and perform Supervised Fine-Tuning (SFT) \cite{zhang2024exploring, cheng2024can, li2025refine}. Both aforementioned types of approaches focus on making the models express \textit{``I know''} or \textit{``I don't know''}. Most aforementioned works find that prompt-based methods are poorly performed. 

We contend that this task is actually challenging for two primary reasons. First, regarding whether a model can itself articulate its own knowledge boundaries, considerable debate persists in current research. For example, \citet{ren2023investigating} states that LLMs struggle to perceive their factual knowledge boundary, and tend to be overconfident, however, \citet{cheng2024can} conclude that the AI assistant can, to a significant extent, identify what it does not know. Second, it is difficult to verify the accuracy of the predicted boundaries.

\subsection{Retrieval-Augmented Generation}
The RAG technique is widely adopted to help models answer certain queries needing external information in both texts \cite{jeong2024adaptive, chen-etal-2024-improving-retrieval, lewis2020retrieval} and image-text scenarios \cite{lin-byrne-2022-retrieval, wu2022multi}. However, current RAG techniques are far from being perfect for enhancing (V)LLMs in all settings. For example, \citet{zhang2024exploring} finds that for math reasoning and code questions, RAG usually brings noise rather than useful information, and thus RAG may even yield adverse effects. Therefore, more effective utilization of RAG can not only result in savings of time and computational resources but also enhance performance in certain scenarios.

\section{Conclusion}

In this paper, we introduce a method with two variants that fine-tunes VLLMs on automatically constructed datasets for boundary identification. This method mitigates the reliance on RAG techniques, which introduce significant latency and long input sequences. Our experiments across diverse held-out VQA datasets, including knowledge-intensive, non-knowledge-intensive, and mixed datasets, demonstrate that our method not only maintains or enhances VLLM performance but also lowers the RAG ratio. Additionally, the fine-tuned knowledge boundary exhibits versatility by functioning as a surrogate for other VLLM series, facilitating retrieval reduction without compromising performance. These findings underscore the efficacy of our method in optimizing the balance between retrieval dependence and model performance, paving the way for more efficient and effective deployment of VLLMs in practical applications.

\section{Limitations}

In this paper, we do not design detailed methods to distinguish the search type, such as text search and image search, towards answering a VQA sample. Experiments utilizing training data sampled from larger VLLMs are currently lacking. Both limitations will be addressed in our future work. Moreover, our current method is not yet able to reliably distinguish errors stemming from visual misrecognition (e.g., incorrect object identification or multi-view ambiguity) from knowledge gaps. 

\section*{Acknowledgment}
This work was supported by Alibaba Group through Alibaba Innovative Research Program and the Core Facility Platform of Computer Science and Communication, SIST, ShanghaiTech University.

\bibliography{custom}
\bibliographystyle{acl_natbib}

\appendix
\clearpage
\newpage

\section{Appendix}
\label{appendix}

\subsection{Training Examples}
\label{training_example_appendix_section}

\paragraph{Hard Knowledge Boundary} Query $\bm{q}$, together with prompts $P_h$ (in blue)\footnote{where <ST\_*> means optional special tokens to specify the position of $\bm{q}$ and indicate the output starting position after <ST\_2>. The detailed format of <ST\_*> and <Image> tokens might need to be modified according to different VL model input formats.}, will be constructed into a training sample $\bm{x}(\bm{q}, P_h)$ as follows:
\begin{framed}
\noindent\textcolor[rgb]{0.392, 0.471, 0.871}{You are an assistant capable of deciding whether a search is needed in a multimodal question-answering scenario. Below, I will provide you with a multimodal question that includes a text question and an image link.
Please respond with "true" or "false," indicating whether a search is necessary (true) or not (false) to answer this multimodal question.
<ST\_1> \\
Text question: \textcolor{black}{$\bm{q_t}$} \\
<Image>: \textcolor{black}{$\bm{q_i}$} \\
<ST\_2>}
\end{framed}

\paragraph{Soft Knowledge Boundary} Query $\bm{q}$, together with prompts $P_s$ (in blue), will be constructed into a training sample $\bm{x}(\bm{q}, P_s)$ as follows: 

\begin{framed}
\noindent\textcolor[rgb]{0.392, 0.471, 0.871}{You are an assistant capable of deciding whether a search is needed in a multimodal question-answering scenario. Below, I will provide you with a multimodal question that includes a text question and an image link.
Please respond with a score ranging from 1.0 to 5.0 indicating whether a search is necessary or not to answer this multimodal question. \\ \\
Follow these guidelines for scoring: \\
- Your score has to be between 1.0 and 5.0, where 1.0 stands for an unnecessary search and 5.0 stands for a necessary search.  \\
- The score does not have to be integer. \\
Example Response: \\
4.0 \\ \\
<ST\_1> \\
Text question: \textcolor{black}{$\bm{q_t}$} \\
<Image>: \textcolor{black}{$\bm{q_i}$} \\
<ST\_2> \\
Your score: }
\end{framed}

\subsection{Training Dataset Description}
\label{training_data_description}

Below is a brief description of each dataset (for training).

\paragraph{InfoSeek} is designed to assess the capability of models to seek and incorporate external information for question answering. It features a variety of queries that necessitate fact retrieval and reasoning that go beyond the provided context.

\paragraph{OK-VQA} is a dataset where images are paired with open-ended questions that require answers stemming from general knowledge that extends beyond the image alone.

\paragraph{VQAv2.0} is a comprehensive VQA dataset that requires interpretation or understanding of the visual content. It features a diverse and balanced range of answers.

\paragraph{MMBench} is a benchmarking suite for evaluating multi-modal understanding, ensuring that multi-modal machine learning systems can effectively process and synthesize data from different sources.

\paragraph{MME} is focused on tasks related to multi-modal entity recognition and extraction. The dataset contains annotations of text and images with multi-modal entities that need to be identified or linked.

\paragraph{Human-Labeled} A group of annotators is asked to annotate whether RAG can help solve a VQA sample. We construct this data to form a reference setting.

\subsection{Training Details and Hyperparameters}
\label{hyper_appendix_section}

\begin{table}[tb]
\centering
\small
\scalebox{1.0}{
\begin{tabular}{ll} 
\toprule
Base Model & Qwen- \& DeepSeek-VL-7B-Chat \\
\midrule
LoRA & $Q,K,V$ \\
\midrule 
LoRA Rank & 8 \\
\midrule
LoRA Alpha & 32 \\
\midrule
Learning Rate & 1e-4 \\ 
\midrule
Optimizer & AdamW \\ 
\midrule
LR Scheduler & Linear \\
\midrule
Precision & bf16 \\
\midrule 
Batch Size & 1 \\
\midrule 
Grad. Accum. & 16 \\
\midrule
GPU & NVIDIA A100-SXM4-80GB \\
\bottomrule
\end{tabular}
}
\caption{Detailed hyperparameters. Grad. Accum. stands for gradient accumulation steps.}
\label{hyper_table}
\end{table}

Recall that our methods need to train a VLLM, parameterized by $\phi$, as a Knowledge Boundary model discussed in Sec.~\ref{training_section}. In experiments, we adopt LoRA \cite{hu2021lora} to optimize $\phi$ and the related hyperparameters are shown in Table~\ref{hyper_table}. We note that our method does not rely heavily on tuning hyperparameters. We just choose intuitive values and it works fairly well. 

\subsection{Training and Evaluation Cost}
Based on our training settings, the training time for both Qwen-VL-Chat-7B and DeepSeek-VL-Chat-7B takes around 10 hours on 1 A100. Besides, the API call costs are as follows:
\begin{enumerate}[leftmargin=*,noitemsep, topsep=1pt]
    \item Training data construction: 339,712 calls to Qwen-Max (for scoring).
    \item Retrieve information from the web: $\sim$3000 calls to Bing/Google (for test set evaluation).
    \item Test set inference: $\sim$3,000 calls to GPT-4o and $\sim$3,000 calls to Qwen-VL-Max
    \item Test set LLM (Qwen-Max) evaluation: $\sim$6,000 calls (No RAG \& All RAG settings)
\end{enumerate}

\subsection{LLM Evaluation Consistency}

We report the LLM evaluation consistency on the Mix dataset with No RAG and All RAG settings in Table~\ref{llm_consistency_table}. We note that the scores of all other settings can be derived from these two settings with the "Use RAG or not" decisions. 
It can be observed that, under these settings, the maximum difference in the average scores between GPT-4o and Qwen-Max is within 3.

\begin{table}[tb]
\centering
\scalebox{0.69}{
\begin{tabular}{l|cc|cc}
\toprule
\textbf{Scoring LLM}      & \textbf{Qwen-Max}  & \textbf{GPT-4o}  & \textbf{Qwen-Max}  & \textbf{GPT-4o}  \\
\textit{Setting}          & \multicolumn{2}{c|}{\textit{No RAG}} & \multicolumn{2}{c}{\textit{All RAG}} \\ 
\midrule
\textbf{Ds.-VL-Chat} & 34.96              & 32.88           & 45.18              & 45.70           \\
\textbf{Qwen-VL-Chat}     & 34.44              & 32.62           & 38.60              & 38.34           \\
\textbf{Qwen-VL-Max}      & 46.54              & 45.88           & 49.26              & 48.45           \\
\textbf{GPT-4o}            & 51.44              & 51.72           & 52.90              & 50.58           \\ 
\bottomrule
\end{tabular}
}
\caption{LLM evaluation consistency between Qwen-Max and GPT-4o on Mix dataset. Scores range from 0 to 100. }
\label{llm_consistency_table}
\end{table}

\subsection{Main Results on DeepSeek}
\label{main_results_ds_section}

We present our main results of DeepSeek-VL-7B-Chat in Table~\ref{main_results_ds_7b_table}. For both the $HKB$ and $SKB$ methods, DeepSeek performs more confidently than Qwen, and it tends to predict a lower ratio of resorting to RAG. On the Mix dataset, DeepSeek also well maintains the performance with the $SKB$ method compared to the All RAG setting and outperforms the Prompt-based method. In addition, compared to Qwen, DeepSeek better utilizes human-labeled data to depict the knowledge boundary and obtains the best result among all settings.

\begin{table*}[t]
    \centering
    \scalebox{0.8}{
    \begin{tabular}{llcccccccc|cc} 
    \toprule
    \textbf{Dataset} & \textbf{Metric} & \begin{tabular}[c]{@{}c@{}}\textbf{No }\\\textbf{RAG}\end{tabular} & \begin{tabular}[c]{@{}c@{}}\textbf{All }\\\textbf{RAG}\end{tabular} & \begin{tabular}[c]{@{}c@{}}\textbf{Prompt-}\\\textbf{based}\end{tabular} & \textbf{\%} & \textbf{HKB} & \textbf{\%} & \textbf{SKB} & \textbf{\%} & \textbf{Human} & \textbf{\%} \\ 
    \midrule
    \multirow{2}{*}{\textbf{Life VQA}} & \textbf{LLM} & 25.81 & 47.35 & 33.36 & 30.20\% & 35.81 & 46.31\% & 42.18 & 73.83\% & 47.21 & 96.64\% \\
     & \textbf{Acc.} & 10.82 & 36.79 & 20.84 & 30.20\% & 24.81 & 46.31\% & 31.93 & 73.83\% & 36.79 & 96.64\% \\ 
    \midrule
    \multirow{2}{*}{\textbf{Private VQA}} & \textbf{LLM} & 22.80 & 27.28 & 23.93 & 21.20\% & 25.45 & 27.60\% & 26.03 & 56.40\% & 27.08 & 88.20\% \\
     & \textbf{Acc.} & 15.51 & 19.75 & 16.38 & 21.20\% & 17.70 & 27.60\% & 17.75 & 56.40\% & 19.57 & 88.20\% \\ 
    \midrule
    \multirow{2}{*}{\textbf{Dyn-VQA ch}} & \textbf{LLM} & 21.32 & 44.20 & 25.63 & 12.48\% & 28.29 & 27.00\% & 37.81 & 79.10\% & 43.54 & 97.42\% \\
     & \textbf{Acc.} & 20.74 & 46.91 & 24.15 & 12.48\% & 28.07 & 27.00\% & 41.18 & 79.10\% & 46.23 & 97.42\% \\ 
    \midrule
    \multirow{2}{*}{\textbf{Dyn-VQA en}} & \textbf{LLM} & 24.90 & 38.36 & 25.63 & 12.73\% & 29.41 & 33.57\% & 32.31 & 60.56\% & 37.77 & 96.78\% \\
     & \textbf{Acc.} & 24.37 & 43.28 & 26.51 & 12.73\% & 30.22 & 33.57\% & 35.49 & 60.56\% & 43.01 & 96.78\% \\ 
    \midrule
    \multirow{2}{*}{\textbf{NoCaps}} & \textbf{LLM} & 63.10 & 59.39 & 62.95 & 2.00\% & 63.12 & 0.20\% & 61.40 & 32.40\% & 62.50 & 6.20\% \\
     & \textbf{Acc.} & 43.89 & 40.45 & 43.62 & 2.00\% & 43.88 & 0.20\% & 42.50 & 32.40\% & 43.48 & 6.20\% \\ 
    \midrule
    \multirow{2}{*}{\textbf{Visual7W}} & \textbf{LLM} & 58.54 & 57.68 & 58.17 & 2.44\% & 58.24 & 7.67\% & 58.16 & 10.98\% & 56.98 & 54.70\% \\
     & \textbf{Acc.} & 46.55 & 46.62 & 46.40 & 2.44\% & 46.27 & 7.67\% & 46.55 & 10.98\% & 46.18 & 54.70\% \\ 
    \midrule
    \midrule
    \multirow{2}{*}{\textbf{Mix}} & \textbf{LLM} & 35.07 & 45.37 & 37.17 & 13.50\% & 39.38 & 25.00\% & 42.46 & 54.83\% & 45.63 & 74.50\% \\
     & \textbf{Acc.} & 25.81 & 35.23 & 28.11 & 13.50\% & 29.08 & 25.00\% & 33.23 & 54.83\% & 35.83 & 74.50\% \\
    \bottomrule
    \end{tabular}
    }
    \caption{Main results of DeepSeek-VL-7B-Chat.}
    \label{main_results_ds_7b_table}
\end{table*}

\subsection{Supplementary Results of ``Surrogate Boundary'' Experiments}
\label{supplementary_results_of_surrogate}

We provide the supplementary experimental results for Sec~\ref{plugin_section} where the token accuracy metrics are shown in Table~\ref{main_results_acc_table}. It can be concluded that similar conclusions can be drawn as in Sec~\ref{plugin_section}. The experiment where DeepSeek-VL-7B-Chat is trained for surrogate boundary prediction is shown in Table~\ref{main_results_ds_llm_table} and \ref{main_results_ds_acc_table}.

\begin{table*}[t]
\centering
\scalebox{0.75}{
\begin{tabular}{llcccccccc|cc} 
\toprule
 & \multicolumn{1}{c}{\begin{tabular}[c]{@{}c@{}}\textbf{Metric:}\\\textbf{Acc.}\end{tabular}} & \begin{tabular}[c]{@{}c@{}}\textbf{No}\\\textbf{RAG}\end{tabular} & \begin{tabular}[c]{@{}c@{}}\textbf{All }\\\textbf{RAG}\end{tabular} & \begin{tabular}[c]{@{}c@{}}\textbf{Prompt-}\\\textbf{based}\end{tabular} & \textbf{\%} & \textbf{HKB} & \textbf{\%} & \textbf{SKB} & \textbf{\%} & \textbf{Human} & \textbf{\%} \\ 
\midrule
\multirow{4}{*}{\textbf{Life VQA}} & Ds.-VL-Chat & 10.82 & 36.79 & 14.12 & 12.75\% & 36.12 & 96.64\% & 30.97 & 61.74\% & 30.50 & 71.14\% \\
 & Qwen-VL-Max & 24.21 & 42.30 & 27.66 & 12.75\% & 41.96 & 96.64\% & 38.37 & 61.74\% & 38.20 & 71.14\% \\
 & Qwen-VL-2 & 23.06 & 41.05 & 27.64 & 12.75\% & 40.71 & 96.64\% & 37.27 & 61.74\% & 37.05 & 71.14\% \\
 & GPT-4o & 31.72 & 40.85 & 32.81 & 12.75\% & 40.85 & 96.64\% & 38.47 & 61.74\% & 41.88 & 71.14\% \\ 
\midrule
\multirow{4}{*}{\textbf{Private VQA}} & Ds.-VL-Chat & 15.51 & 19.75 & 16.65 & 14.80\% & 19.75 & 99.20\% & 18.20 & 67.80\% & 18.51 & 72.00\% \\
 & Qwen-VL-Max & 27.93 & 28.14 & 28.08 & 14.80\% & 28.29 & 99.20\% & 27.68 & 67.80\% & 28.96 & 72.00\% \\
 & Qwen-VL-2 & 27.69 & 30.72 & 27.75 & 14.80\% & 30.87 & 99.20\% & 28.96 & 67.80\% & 31.13 & 72.00\% \\
 & GPT-4o & 31.12 & 27.02 & 30.88 & 14.80\% & 26.87 & 99.20\% & 27.72 & 67.80\% & 29.10 & 72.00\% \\ 
\midrule
\multirow{4}{*}{\textbf{Dyn-VQA ch}} & Ds.-VL-Chat & 20.74 & 46.91 & 22.37 & 6.38\% & 46.05 & 95.66\% & 44.13 & 84.26\% & 33.60 & 46.95\% \\
 & Qwen-VL-Max & 31.53 & 46.73 & 33.53 & 6.38\% & 46.38 & 95.66\% & 44.82 & 84.26\% & 39.79 & 46.95\% \\
 & Qwen-VL-2 & 31.52 & 46.70 & 33.52 & 6.38\% & 46.28 & 95.66\% & 44.69 & 84.26\% & 39.85 & 46.95\% \\
 & GPT-4o & 36.46 & 51.27 & 37.32 & 6.38\% & 50.85 & 95.66\% & 49.4 & 84.26\% & 42.45 & 46.95\% \\ 
\midrule
\multirow{4}{*}{\textbf{Dyn-VQA en}} & Ds.-VL-Chat & 24.37 & 43.28 & 26.80 & 14.13\% & 42.08 & 89.79\% & 40.61 & 76.08\% & 31.67 & 29.51\% \\
 & Qwen-VL-Max & 37.54 & 45.27 & 38.03 & 14.13\% & 44.30 & 89.79\% & 43.55 & 76.08\% & 39.40 & 29.51\% \\
 & Qwen-VL-2 & 37.37 & 45.16 & 37.25 & 14.13\% & 43.84 & 89.79\% & 43.48 & 76.08\% & 40.66 & 29.51\% \\
 & GPT-4o & 43.33 & 49.71 & 42.40 & 14.13\% & 48.48 & 89.79\% & 47.66 & 76.08\% & 45.07 & 29.51\% \\ 
\midrule
\multirow{4}{*}{\textbf{NoCaps}} & Ds.-VL-Chat & 43.89 & 40.45 & 43.89 & 0.00\% & 42.76 & 38.40\% & 43.89 & 0.00\% & 43.89 & 0.00\% \\
 & Qwen-VL-Max & 37.47 & 34.55 & 37.47 & 0.00\% & 36.75 & 38.40\% & 37.47 & 0.00\% & 37.47 & 0.00\% \\
 & Qwen-VL-2 & 37.26 & 34.61 & 37.26 & 0.00\% & 36.35 & 38.40\% & 37.26 & 0.00\% & 37.26 & 0.00\% \\
 & GPT-4o & 32.12 & 36.25 & 32.12 & 0.00\% & 33.22 & 38.40\% & 32.12 & 0.00\% & 32.12 & 0.00\% \\ 
\midrule
\multirow{4}{*}{\textbf{Visual7W}} & Ds.-VL-Chat & 46.55 & 46.62 & 46.29 & 31.36\% & 46.03 & 35.37\% & 46.58 & 2.96\% & 46.55 & 0.52\% \\
 & Qwen-VL-Max & 46.07 & 44.44 & 48.63 & 31.36\% & 45.16 & 35.37\% & 46.13 & 2.96\% & 46.07 & 0.52\% \\
 & Qwen-VL-2 & 45.94 & 43.86 & 48.47 & 31.36\% & 45.06 & 35.37\% & 45.99 & 2.96\% & 45.94 & 0.52\% \\
 & GPT-4o & 41.59 & 37.16 & 40.09 & 31.36\% & 39.41 & 35.37\% & 41.80 & 2.96\% & 41.48 & 0.52\% \\ 
\midrule
\midrule
\multirow{4}{*}{\textbf{Mix}} & Ds.-VL-Chat & 25.81 & 35.23 & 26.55 & 12.67\% & \textbf{35.38} & 76.83\% & 33.06 & 49.33\% & 32.73 & 38.33\% \\
 & Qwen-VL-Max & 32.35 & 34.78 & 33.00 & 12.67\% & \uline{35.48} & 76.83\% & \uline{34.84} & 49.33\% & 35.51 & 38.33\% \\
 & Qwen-VL-2 & 32.59 & 35.56 & 33.27 & 12.67\% & \uline{36.29} & 76.83\% & \uline{35.62} & 49.33\% & 36.33 & 38.33\% \\
 & GPT-4o & 34.52 & 35.96 & 33.99 & 12.67\% & 35.90 & 76.83\% & 35.86 & 49.33\% & 36.49 & 38.33\% \\
\bottomrule
\end{tabular}
}
\caption{Knowledge Boundary model (Qwen-VL-7B-Chat) as a surrogate boundary identifier for other VLLMs. Results evaluated by token accuracy.}
\label{main_results_acc_table}
\end{table*}

\begin{table*}[t]
\centering
\scalebox{0.75}{
\begin{tabular}{llcccccccc|cc} 
\toprule
 & \multicolumn{1}{c}{\begin{tabular}[c]{@{}c@{}}\textbf{Metric:}\\\textbf{LLM}\end{tabular}} & \begin{tabular}[c]{@{}c@{}}\textbf{No}\\\textbf{RAG}\end{tabular} & \begin{tabular}[c]{@{}c@{}}\textbf{All }\\\textbf{RAG}\end{tabular} & \begin{tabular}[c]{@{}c@{}}\textbf{Prompt-}\\\textbf{based}\end{tabular} & \textbf{\%} & \textbf{HKB} & \textbf{\%} & \textbf{SKB} & \textbf{\%} & \textbf{Human} & \textbf{\%} \\ 
\midrule
\multirow{4}{*}{\textbf{Life VQA}} & Qwen-VL-Chat & 31.14 & 42.85 & 33.62 & 30.20\% & 37.08 & 46.31\% & 41.78 & 73.83\% & 43.05 & 96.64\% \\
 & Qwen-VL-Max & 44.09 & 56.64 & 46.51 & 30.20\% & 48.59 & 46.31\% & 54.16 & 73.83\% & 56.51 & 96.64\% \\
 & Qwen-VL-2 & 42.95 & 54.23 & 45.00 & 30.20\% & 48.66 & 46.31\% & 53.36 & 73.83\% & 54.23 & 96.64\% \\
 & GPT-4o & 47.45 & 56.38 & 53.15 & 30.20\% & 54.16 & 46.31\% & 55.37 & 73.83\% & 56.11 & 96.64\% \\ 
\midrule
\multirow{4}{*}{\textbf{Private VQA}} & Qwen-VL-Chat & 24.45 & 26.16 & 25.21 & 21.20\% & 25.35 & 27.60\% & 26.08 & 56.40\% & 26.01 & 88.20\% \\
 & Qwen-VL-Max & 36.84 & 42.97 & 36.45 & 21.20\% & 39.16 & 27.60\% & 42.26 & 56.40\% & 43.73 & 88.20\% \\
 & Qwen-VL-2 & 36.76 & 38.20 & 36.65 & 21.20\% & 38.52 & 27.60\% & 39.37 & 56.40\% & 39.03 & 88.20\% \\
 & GPT-4o & 40.13 & 38.72 & 39.15 & 21.20\% & 40.59 & 27.60\% & 40.76 & 56.40\% & 39.56 & 88.20\% \\ 
\midrule
\multirow{4}{*}{\textbf{Dyn-VQA ch}} & Qwen-VL-Chat & 37.73 & 44.68 & 38.44 & 12.48\% & 40.31 & 27.00\% & 39.63 & 79.10\% & 43.85 & 97.42\% \\
 & Qwen-VL-Max & 32.67 & 50.85 & 35.11 & 12.48\% & 37.20 & 27.00\% & 46.62 & 79.10\% & 50.32 & 97.42\% \\
 & Qwen-VL-2 & 45.95 & 50.91 & 46.33 & 12.48\% & 48.03 & 27.00\% & 46.42 & 79.10\% & 50.13 & 97.42\% \\
 & GPT-4o & 42.10 & 56.51 & 44.55 & 12.48\% & 44.80 & 27.00\% & 51.99 & 79.10\% & 56.23 & 97.42\% \\ 
\midrule
\multirow{4}{*}{\textbf{Dyn-VQA en}} & Qwen-VL-Chat & 22.07 & 35.23 & 23.58 & 12.73\% & 26.91 & 33.57\% & 30.06 & 60.56\% & 34.27 & 96.78\% \\
 & Qwen-VL-Max & 19.41 & 39.90 & 22.86 & 12.73\% & 24.71 & 33.57\% & 36.01 & 60.56\% & 39.44 & 96.78\% \\
 & Qwen-VL-2 & 37.90 & 44.29 & 38.58 & 12.73\% & 40.45 & 33.57\% & 40.11 & 60.56\% & 43.58 & 96.78\% \\
 & GPT-4o & 32.73 & 51.17 & 35.25 & 12.73\% & 37.36 & 33.57\% & 47.15 & 60.56\% & 50.65 & 96.78\% \\ 
\midrule
\multirow{4}{*}{\textbf{NoCaps}} & Qwen-VL-Chat & 50.46 & 30.41 & 50.00 & 2.00\% & 50.48 & 0.20\% & 44.43 & 32.40\% & 49.48 & 6.20\% \\
 & Qwen-VL-Max & 62.04 & 49.63 & 61.82 & 2.00\% & 61.92 & 0.20\% & 57.63 & 32.40\% & 61.16 & 6.20\% \\
 & Qwen-VL-2 & 61.88 & 49.84 & 61.66 & 2.00\% & 61.78 & 0.20\% & 57.44 & 32.40\% & 60.92 & 6.20\% \\
 & GPT-4o & 61.58 & 64.51 & 61.68 & 2.00\% & 61.56 & 0.20\% & 62.00 & 32.40\% & 61.68 & 6.20\% \\ 
\midrule
\multirow{4}{*}{\textbf{Visual7W}} & Qwen-VL-Chat & 55.53 & 54.52 & 55.45 & 2.44\% & 55.76 & 7.67\% & 55.31 & 10.98\% & 55.01 & 54.70\% \\
 & Qwen-VL-Max & 61.72 & 58.16 & 61.59 & 2.44\% & 61.27 & 7.67\% & 61.08 & 10.98\% & 59.04 & 54.70\% \\
 & Qwen-VL-2 & 61.81 & 58.07 & 61.58 & 2.44\% & 61.25 & 7.67\% & 61.23 & 10.98\% & 58.78 & 54.70\% \\
 & GPT-4o & 53.34 & 47.44 & 53.30 & 2.44\% & 52.71 & 7.67\% & 52.70 & 10.98\% & 49.97 & 54.70\% \\ 
\midrule
\midrule
\multirow{4}{*}{\textbf{Mix}} & Qwen-VL-Chat & 34.58 & 39.03 & 35.54 & 13.50\% & 37.12 & 25.00\% & \uline{39.76} & 54.83\% & 42.61 & 74.50\% \\
 & Qwen-VL-Max & 46.13 & 49.02 & 46.47 & 13.50\% & 47.43 & 25.00\% & 48.98 & 54.83\% & 51.39 & 74.50\% \\
 & Qwen-VL-2 & 46.26 & 47.84 & 46.64 & 13.50\% & \uline{48.17} & 25.00\% & \uline{48.55} & 54.83\% & 50.13 & 74.50\% \\
 & GPT-4o & 51.21 & 52.70 & 51.81 & 13.50\% & 52.28 & 25.00\% & 52.04 & 54.83\% & 53.42 & 74.50\% \\
\bottomrule
\end{tabular}
}
\caption{Knowledge Boundary model (DeepSeek-VL-7B-Chat) as a surrogate boundary identifier for other VLLMs. Results evaluated by LLM.}
\label{main_results_ds_llm_table}
\end{table*}

\begin{table*}[t]
\centering
\scalebox{0.75}{
\begin{tabular}{llcccccccc|cc} 
\toprule
 & \multicolumn{1}{c}{\begin{tabular}[c]{@{}c@{}}\textbf{Metric:}\\\textbf{Acc.}\end{tabular}} & \begin{tabular}[c]{@{}c@{}}\textbf{No}\\\textbf{RAG}\end{tabular} & \begin{tabular}[c]{@{}c@{}}\textbf{All }\\\textbf{RAG}\end{tabular} & \begin{tabular}[c]{@{}c@{}}\textbf{Prompt-}\\\textbf{based}\end{tabular} & \textbf{\%} & \textbf{HKB} & \textbf{\%} & \textbf{SKB} & \textbf{\%} & \textbf{Human} & \textbf{\%} \\ 
\midrule
\multirow{4}{*}{\textbf{Life VQA}} & Qwen-VL-Chat & 17.80 & 36.11 & 23.68 & 30.20\% & 28.05 & 46.31\% & 34.43 & 73.83\% & 36.78 & 96.64\% \\
 & Qwen-VL-Max & 25.42 & 42.30 & 30.09 & 30.20\% & 32.83 & 46.31\% & 38.38 & 73.83\% & 42.07 & 96.64\% \\
 & Qwen-VL-2 & 25.29 & 41.05 & 29.77 & 30.20\% & 33.75 & 46.31\% & 38.48 & 73.83\% & 40.83 & 96.64\% \\
 & GPT-4o & 31.72 & 40.85 & 36.49 & 30.20\% & 38.53 & 46.31\% & 40.01 & 73.83\% & 42.19 & 96.64\% \\ 
\midrule
\multirow{4}{*}{\textbf{Private VQA}} & Qwen-VL-Chat & 16.26 & 18.40 & 17.28 & 21.20\% & 18.11 & 27.60\% & 18.34 & 56.40\% & 18.90 & 88.20\% \\
 & Qwen-VL-Max & 27.12 & 28.14 & 26.77 & 21.20\% & 27.94 & 27.60\% & 28.18 & 56.40\% & 28.31 & 88.20\% \\
 & Qwen-VL-2 & 27.04 & 30.72 & 27.89 & 21.20\% & 28.78 & 27.60\% & 29.94 & 56.40\% & 30.95 & 88.20\% \\
 & GPT-4o & 31.12 & 27.02 & 29.74 & 21.20\% & 30.78 & 27.60\% & 29.73 & 56.40\% & 28.24 & 88.20\% \\ 
\midrule
\multirow{4}{*}{\textbf{Dyn-VQA ch}} & Qwen-VL-Chat & 37.37 & 45.16 & 38.25 & 12.48\% & 39.83 & 27.00\% & 39.37 & 79.10\% & 44.84 & 97.42\% \\
 & Qwen-VL-Max & 31.66 & 46.70 & 34.29 & 12.48\% & 35.11 & 27.00\% & 42.80 & 79.10\% & 46.35 & 97.42\% \\
 & Qwen-VL-2 & 43.33 & 49.71 & 43.68 & 12.48\% & 45.47 & 27.00\% & 45.17 & 79.10\% & 49.28 & 97.42\% \\
 & GPT-4o & 36.46 & 51.27 & 38.75 & 12.48\% & 39.78 & 27.00\% & 46.96 & 79.10\% & 51.13 & 97.42\% \\ 
\midrule
\multirow{4}{*}{\textbf{Dyn-VQA en}} & Qwen-VL-Chat & 25.64 & 41.87 & 27.33 & 12.73\% & 31.66 & 33.57\% & 35.00 & 60.56\% & 41.63 & 96.78\% \\
 & Qwen-VL-Max & 23.41 & 43.06 & 26.81 & 12.73\% & 29.04 & 33.57\% & 39.36 & 60.56\% & 42.57 & 96.78\% \\
 & Qwen-VL-2 & 37.54 & 45.27 & 37.52 & 12.73\% & 40.05 & 33.57\% & 40.50 & 60.56\% & 44.95 & 96.78\% \\
 & GPT-4o & 31.66 & 46.73 & 34.25 & 12.73\% & 34.93 & 33.57\% & 42.96 & 60.56\% & 46.39 & 96.78\% \\ 
\midrule
\multirow{4}{*}{\textbf{NoCaps}} & Qwen-VL-Chat & 40.50 & 30.72 & 40.39 & 2.00\% & 40.49 & 0.20\% & 37.88 & 32.40\% & 39.92 & 6.20\% \\
 & Qwen-VL-Max & 37.47 & 34.55 & 37.44 & 2.00\% & 37.42 & 0.20\% & 36.47 & 32.40\% & 37.22 & 6.20\% \\
 & Qwen-VL-2 & 37.26 & 34.61 & 37.30 & 2.00\% & 37.21 & 0.20\% & 36.21 & 32.40\% & 37.04 & 6.20\% \\
 & GPT-4o & 32.12 & 36.25 & 32.23 & 2.00\% & 32.12 & 0.20\% & 32.96 & 32.40\% & 32.35 & 6.20\% \\ 
\midrule
\multirow{4}{*}{\textbf{Visual7W}} & Qwen-VL-Chat & 44.34 & 44.94 & 44.26 & 2.44\% & 44.64 & 7.67\% & 44.86 & 10.98\% & 45.11 & 54.70\% \\
 & Qwen-VL-Max & 49.41 & 45.13 & 49.39 & 2.44\% & 49.28 & 7.67\% & 48.13 & 10.98\% & 46.04 & 54.70\% \\
 & Qwen-VL-2 & 49.71 & 44.19 & 49.48 & 2.44\% & 49.58 & 7.67\% & 48.43 & 10.98\% & 45.51 & 54.70\% \\
 & GPT-4o & 41.59 & 37.16 & 41.76 & 2.44\% & 40.96 & 7.67\% & 40.91 & 10.98\% & 39.10 & 54.70\% \\ 
\midrule
\midrule
\multirow{4}{*}{\textbf{Mix}} & Qwen-VL-Chat & 26.13 & 32.39 & 28.00 & 13.50\% & 29.55 & 25.00\% & \uline{32.46} & 54.83\% & 34.06 & 74.50\% \\
 & Qwen-VL-Max & 32.35 & 34.78 & 32.91 & 13.50\% & 33.17 & 25.00\% & \uline{35.12} & 54.83\% & 35.96 & 74.50\% \\
 & Qwen-VL-2 & 32.45 & 35.56 & 33.51 & 13.50\% & 33.86 & 25.00\% & \uline{35.88} & 54.83\% & 36.63 & 74.50\% \\
 & GPT-4o & 34.52 & 35.96 & 35.17 & 13.50\% & 35.77 & 25.00\% & 35.86 & 54.83\% & 36.24 & 74.50\% \\
\bottomrule
\end{tabular}
}
\caption{Knowledge Boundary model (DeepSeek-VL-7B-Chat) as a surrogate boundary identifier for other VLLMs. Results were evaluated by token accuracy.}
\label{main_results_ds_acc_table}
\end{table*}

\subsection{Supplementary Results on MMMU Dataset}
\label{mmmu_appendix_section}

In this section, we show the experimental results of our methods on a challenging dataset, MMMU\footnote{We converted the dataset's original multiple-choice format into a conventional VQA format to ensure consistency with the aforementioned experimental settings.} \cite{yue2023mmmu} in Table~\ref{mmmu_results_llm_table}. MMMU is a dataset containing VQA samples demanding college-level subject knowledge and deliberate reasoning, and it is hard to verify the knowledge boundary that our methods depict by simply adopting RAG.  

\begin{figure*}[tb]
    \centering
    \includegraphics[width=0.49\linewidth]{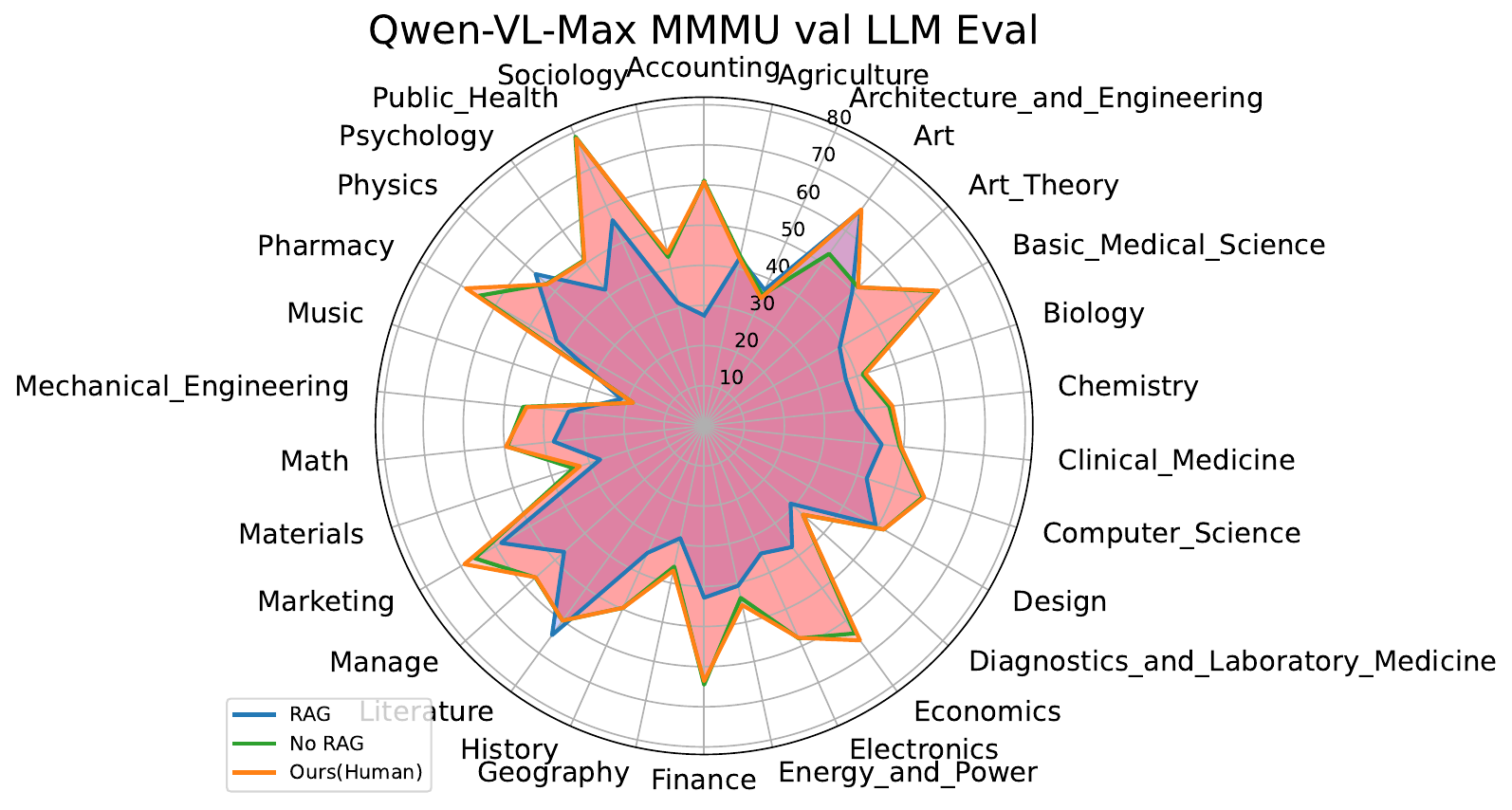}
    \includegraphics[width=0.49\linewidth]{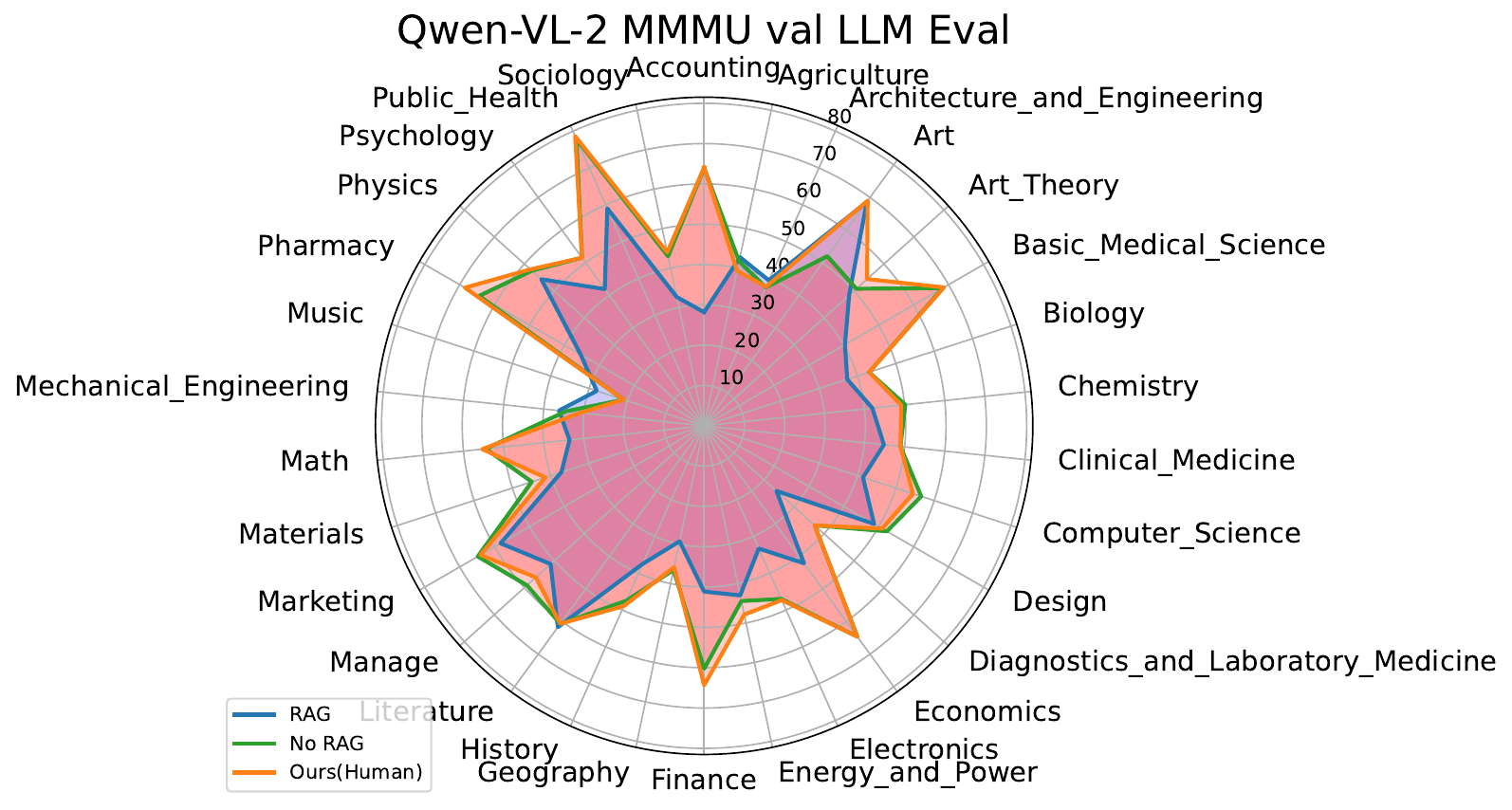}
    \includegraphics[width=0.49\linewidth]{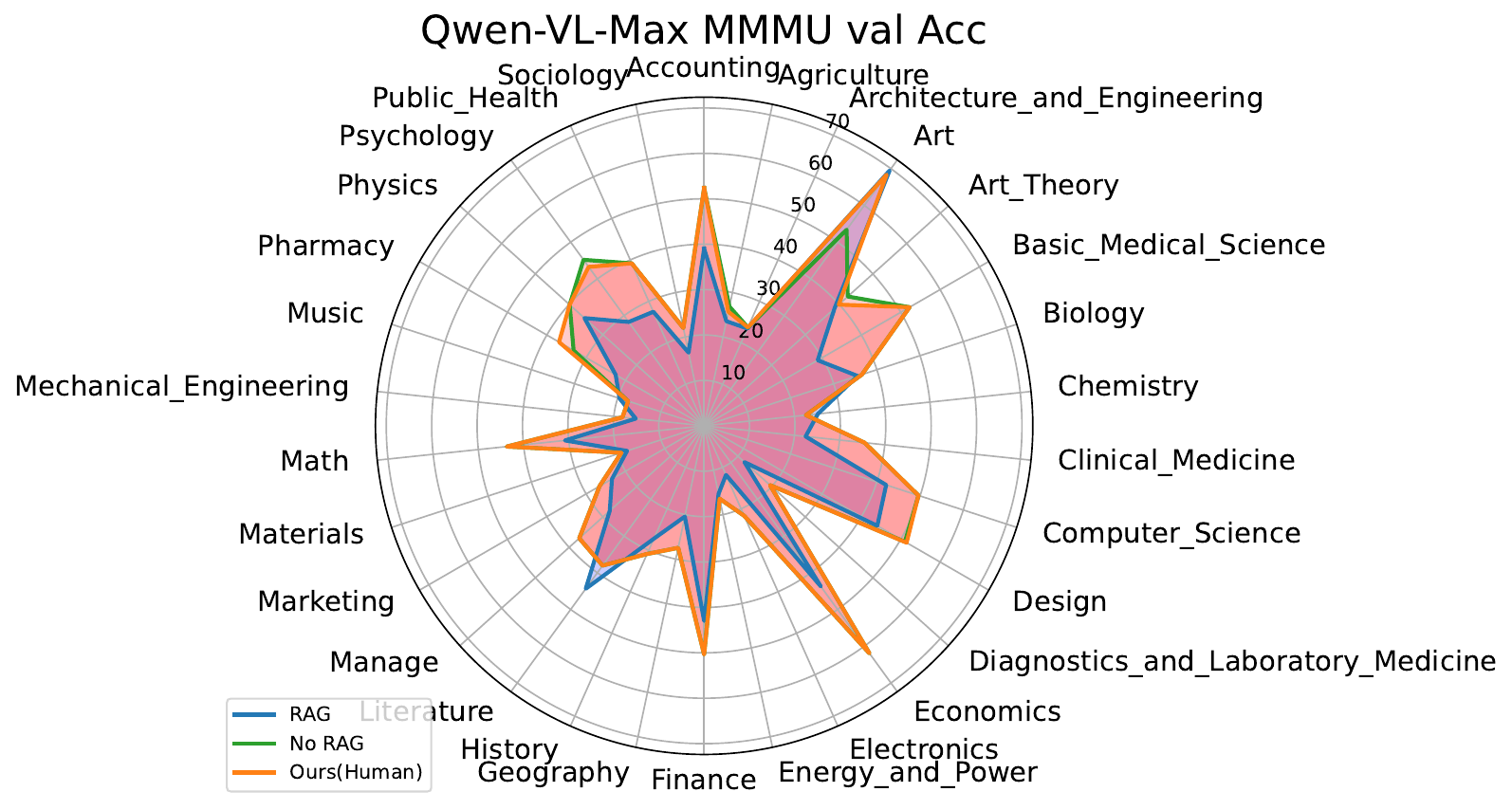}
    \includegraphics[width=0.49\linewidth]{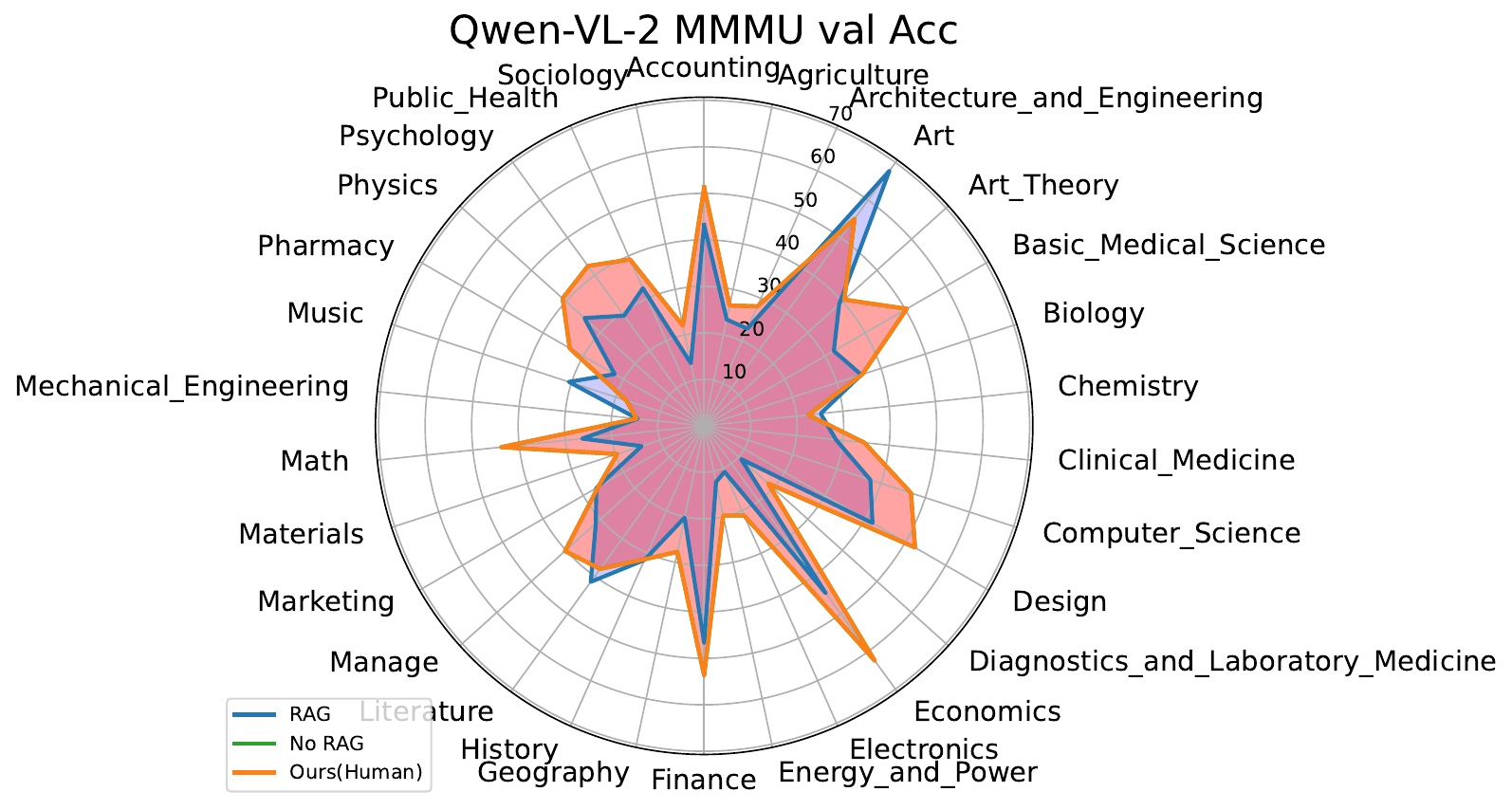}
    \caption{Qwen-VL-Max and Qwen-VL-2 performance on the MMMU validation set with the Knowledge Boundary model trained on Human-labeled data.}
    \label{mmmu_radar_fig}
\end{figure*}

The results in Table~\ref{mmmu_results_llm_table} show that the Knowledge Boundary model trained by human-labeled data helps achieve the best performance. It verifies that the aforementioned Human-labeled training data is effective. In addition, we show that our methods also exhibit substantial potential within this setting, in which both the $HKB$ and $SKB$ models predict a high search ratio over MMMU. We contend that the suboptimal performance of this dataset arises because it lies beyond the knowledge boundaries, which are challenging to validate using RAG, as delineated by the white dashed lines in Fig.~\ref{outline}. We present the performance of each of the 30 subjects in the MMMU validation set in Fig~\ref{mmmu_radar_fig}. The first row shows the LLM evaluation results, and the second shows the token accuracy metric. We can see that in most subjects ``Human'' setting succeeds in obtaining a higher performance than both ``All RAG'' and ``No RAG'' settings.

\begin{table*}
\centering
\scalebox{0.85}{
\begin{tabular}{llcccccccc} 
\toprule
 &  & \textbf{No RAG} & \textbf{All RAG} & \textbf{Human} & \textbf{\%} & \textbf{HKB} & \textbf{\%} & \textbf{SKB} & \textbf{\%} \\ 
\midrule
\multirow{4}{*}{\textbf{MMMU}} & Qwen-VL-Chat & 20.12 & 20.28 & \textbf{21.24} & 6.88\% & \uline{20.35} & 97.08\% & 20.18 & 61.26\% \\
 & Qwen-VL-Max & 51.33 & 41.37 & \textbf{52.67} & 6.88\% & 41.46 & 97.08\% & 44.40 & 61.26\% \\
 & Qwen-VL-2 & 51.45 & 42.39 & \textbf{51.93} & 6.88\% & 42.54 & 97.08\% & 45.61 & 61.26\% \\
 & GPT-4o & 56.60 & 56.64 & \textbf{57.36} & 6.88\% & \uline{56.92} & 97.08\% & \uline{56.91} & 61.26\% \\
\bottomrule
\end{tabular}
}
\caption{Results evaluated by LLM on MMMU validation set.}
\label{mmmu_results_llm_table}
\end{table*}

\end{document}